\documentclass[a4paper,fleqn]{cas-dc}

\usepackage[authoryear]{natbib}
\usepackage{adjustbox}
\usepackage{amsmath,amssymb,amsfonts}

\usepackage{graphicx}
\usepackage{textcomp}
\usepackage{booktabs}
\usepackage{caption}
\usepackage{rotating}
\usepackage{multirow}
\usepackage{booktabs}
\usepackage{adjustbox}
\usepackage{pifont}
\usepackage{algpseudocode}
\usepackage{amsmath}
\usepackage{amsfonts}
\usepackage[ruled,vlined]{algorithm2e}
\usepackage{amsmath}
\SetInd{1em}{1.5em}

\def\tsc#1{\csdef{#1}{\textsc{\lowercase{#1}}\xspace}}
\tsc{WGM}
\tsc{QE}
\tsc{EP}
\tsc{PMS}
\tsc{BEC}
\tsc{DE}
\begin{document}

\let\WriteBookmarks\relax
\def\floatpagepagefraction{1}
\def\textpagefraction{.001}

\shorttitle{FSDA-DG}

\shortauthors{Zanting Ye et~al.}

\title [mode = title]{FSDA-DG: Improving Cross-Domain Generalizability 
of Medical Image Segmentation with Few Source Domain Annotations}                     

\tnotetext[1]{Corresponding authors.}

%\cormark[1]

\author[1]{Zanting Ye}[auid=000,bioid=1,
                    orcid=0009-0006-8874-8882]

\affiliation[1]{organization={School of Biomedical Engineering},
    addressline={Southern Medical University}, 
    city={Guangzhou},
    % citysep={}, % Uncomment if no comma needed between city and postcode
    %postcode={1043 NX}, 
    state={Guangdong},
    postcode={510515}, 
    country={China}}
\author[2,3]{Ke Wang}


%\ead[url]{www.cvr.cc, cvr@sayahna.org}

%\credit{Conceptualization of this study, Methodology, Software}

\affiliation[2]{organization={College of Information Science and Technology},
    addressline={Zhejiang Shuren University}, 
    city={Hangzhou},
    state={Zhejiang},
    postcode={310015},
    % citysep={}, % Uncomment if no comma needed between city and postcode
    %postcode={1043 NX}, 
    country={China}}

\affiliation[3]{organization={State Key Laboratory of Industrial Control Technology},
    addressline={Zhejiang University}, 
    city={Hangzhou},
     state={Zhejiang},
    postcode={310058},
    % citysep={}, % Uncomment if no comma needed between city and postcode
    %postcode={1043 NX}, 
    country={China}}

\author[4]{Wenbing Lv}
%\cormark[2]
%\fnmark[1,3]
%\ead{Xiang.Xie@newcastle.ac.uk}
%\ead[URL]{www.stmdocs.in}

\affiliation[4]{organization={School of Information and Yunnan Key Laboratory of Intelligent Systems and Computing},
    addressline={Yunnan University}, 
    city={Kunming},
    state={Yunnan},
    postcode={650091}, 
    % citysep={}, % Uncomment if no comma needed between city and postcode
    %postcode={695571}, 
    %state={Newcastle upon Tyne},
    country={China}}

\author[1,5,6]{Qianjin Feng}
%\cormark[2]
%\fnmark[1,3]
%\ead{Xiang.Xie@newcastle.ac.uk}
%\ead[URL]{www.stmdocs.in}

\affiliation[5]{organization={Guangdong Provincial Key Laboratory of Medical Image Processing},
    addressline={Southern Medical University}, 
    city={Guangzhou},
    state={Guangdong},
    postcode={510515}, 
    % citysep={}, % Uncomment if no comma needed between city and postcode
    %postcode={695571}, 
    %state={Newcastle upon Tyne},
    country={China}}
    
\affiliation[6]{organization={Guangdong Province Engineering Laboratory for Medical Imaging and Diagnostic Technology},
    addressline={Southern Medical University}, 
    city={Guangzhou},
    state={Guangdong},
    postcode={510515}, 
    % citysep={}, % Uncomment if no comma needed between city and postcode
    %postcode={695571}, 
    %state={Newcastle upon Tyne},
    country={China}}
    
\author[1,5,6,7]{Lijun Lu}
%\cormark[2]
%\fnmark[1,3]
%\ead{Xiang.Xie@newcastle.ac.uk}
%\ead[URL]{www.stmdocs.in}

\affiliation[7]{organization={Pazhou Lab}, 
    city={Guangzhou},
    state={Guangdong},
    postcode={510320}, 
    % citysep={}, % Uncomment if no comma needed between city and postcode
    %postcode={695571}, 
    %state={Newcastle upon Tyne},
    country={China}}
\cormark[1]
\ead{ljlubme@gmail.com}


%\ead[URL]{www.sayahna.org}

%\credit{Data curation, Writing - Original draft preparation}

%\cortext{E-mail address: 3110102872@zju.edu.cn(B.Liu)}

%\cortext[cor2]{Principal corresponding author}

%\fntext[fn1]{This is the first author footnote. but is common to third
%  author as well.}
%\fntext[fn2]{Another author footnote, this is a very long footnote and
%  it should be a really long footnote. But this footnote is not yet
%  sufficiently long enough to make two lines of footnote text.}

%\nonumnote{This note has no numbers. In this work we demonstrate $a_b$
%  the formation Y\_1 of a new type of polariton on the interface
%  between a cuprous oxide slab and a polystyrene micro-sphere placed
%  on the slab.
 % }


\begin{abstract}
Deep learning-based medical image segmentation faces significant challenges arising from limited labeled data and domain shifts. While prior approaches have primarily addressed these issues independently, their simultaneous occurrence is common in medical imaging. A method that generalizes to unseen domains using only minimal annotations offers significant practical value due to reduced data annotation and development costs. In pursuit of this goal, we propose FSDA-DG, a novel solution to improve cross-domain generalizability of medical image segmentation with few single-source domain annotations. Specifically, our approach introduces semantics-guided semi-supervised data augmentation. This method divides images into global broad regions and semantics-guided local regions, and applies distinct augmentation strategies to enrich data distribution. Within this framework, both labeled and unlabeled data are transformed into extensive domain knowledge while preserving domain-invariant semantic information. Additionally, FSDA-DG employs a multi-decoder U-Net pipeline semi-supervised learning (SSL) network to improve domain-invariant representation learning through consistent prior assumption across multiple perturbations. By integrating data-level and model-level designs, FSDA-DG achieves superior performance compared to state-of-the-art methods in two challenging single domain generalization (SDG) tasks with limited annotations. The code is publicly available at https://github.com/yezanting/FSDA-DG.
\end{abstract}

% \includegraphics{figs/grabs.pdf}

% \begin{highlights}
% \item We propose a pioneering framework, MLN-net, devised explicitly for the segmentation of clustered microcalcifications. MLN-net is proficient in accurately segmenting multi-source images using single-source images, offering a substantial solution to the domain shift challenge between source and target domains.
% \item We incorporate a groundbreaking segmentation network with multiple LN layers. This network exhibits a superior ability to capture the distinctive features inherent in multi-source images.
% \item We pioneers a source domain data augmentation method leveraging Bézier curves and grayscale-inversion transformation, thereby significantly enhancing the diversity of source domain data. Moreover, we develop a branch selection strategy to measure the similarities between the source domain data and the target domain data. 
% \item To validate the efficacy of the proposed MLN-net, comprehensive analyses, including ablation studies, are carried out alongside a comparison with 12 established baseline methods.

% \end{highlights}

% Keywords

\maketitle

\begin{figure}[!t]
\centerline{\includegraphics[width=1\linewidth]{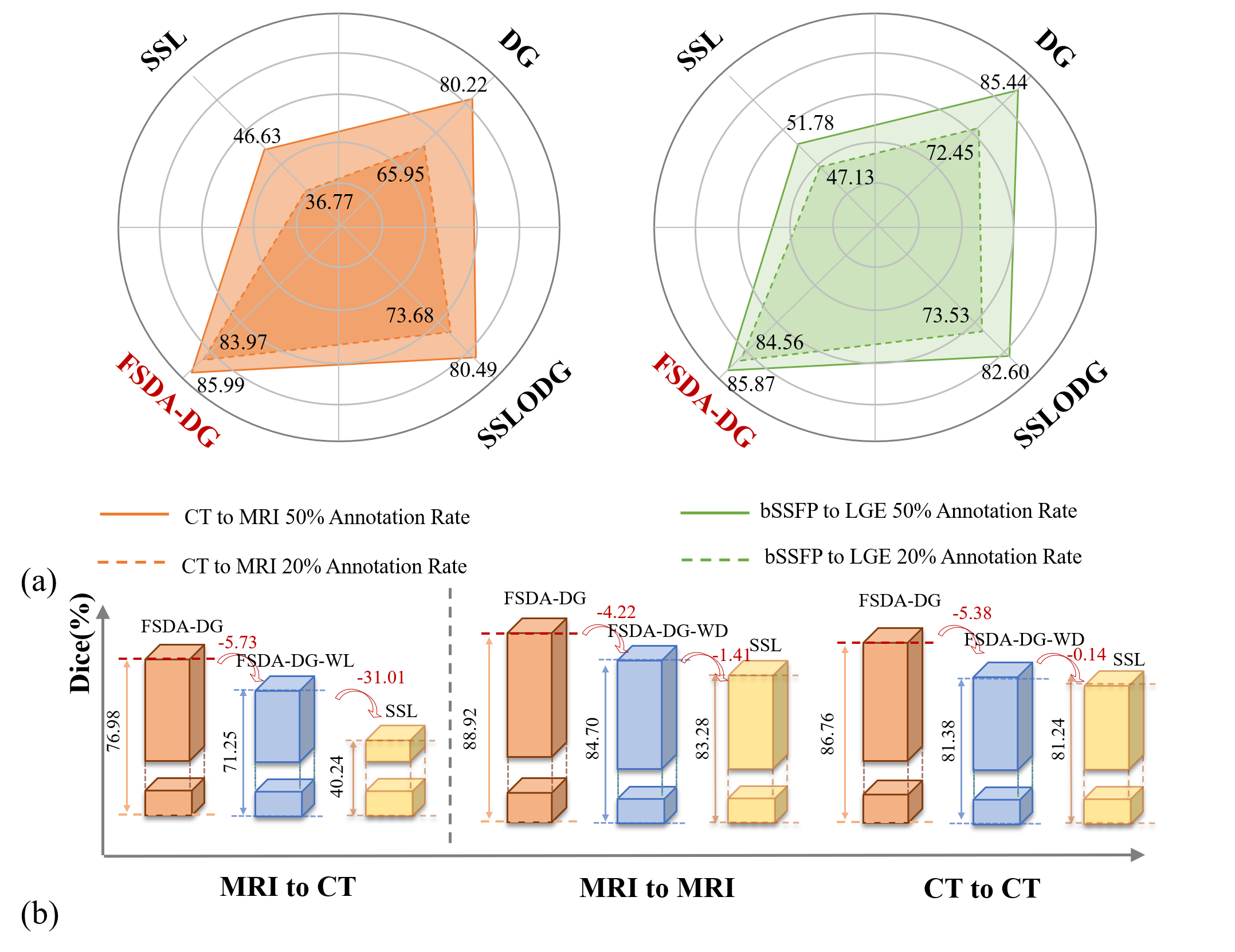}}
\caption{(a) Performance comparison of various methods under modality domain shifts. (b) Comparison of method performance for intra-dataset versus modality domain shifts at a 20\% labeling rate. FSDA-DG-WL and FSDA-DG-WD denote ablation versions of our model without the multi-decoder U-Net pipeline SSL network and the semantics-guided semi-supervised data augmentation, respectively.}
\label{fig1}
\end{figure}

\section{Introduction}
Deep learning-based medical image segmentation \cite{hering2022learn2reg,dalmaz2022resvit,sourati2019intelligent}, a fundamental and critical step in computer-aided diagnosis systems, enhances diagnostic efficiency \cite{you2022simcvd,selvan2022carbon}. Despite its success \cite{haenssle2018man,liu2019comparison,aljabri2022review,ronneberger2015u}, most deep learning methods rely heavily on high-quality data and precise annotations. This dependency presents significant challenges stemming from the sensitive nature of medical data, its limited shareability, and the high costs of obtaining pixel- or voxel-level annotations \cite{luo2022semi,qureshi2022medical}. Furthermore, domain shifts, caused by variations in scanners, scanner generations, and imaging protocols, introduce another layer of complexity by altering the data distribution of medical images. These challenges hinder the widespread adoption of deep learning in clinical workflows.

\begin{figure}[t]
\centering
\includegraphics[width=0.47\textwidth]{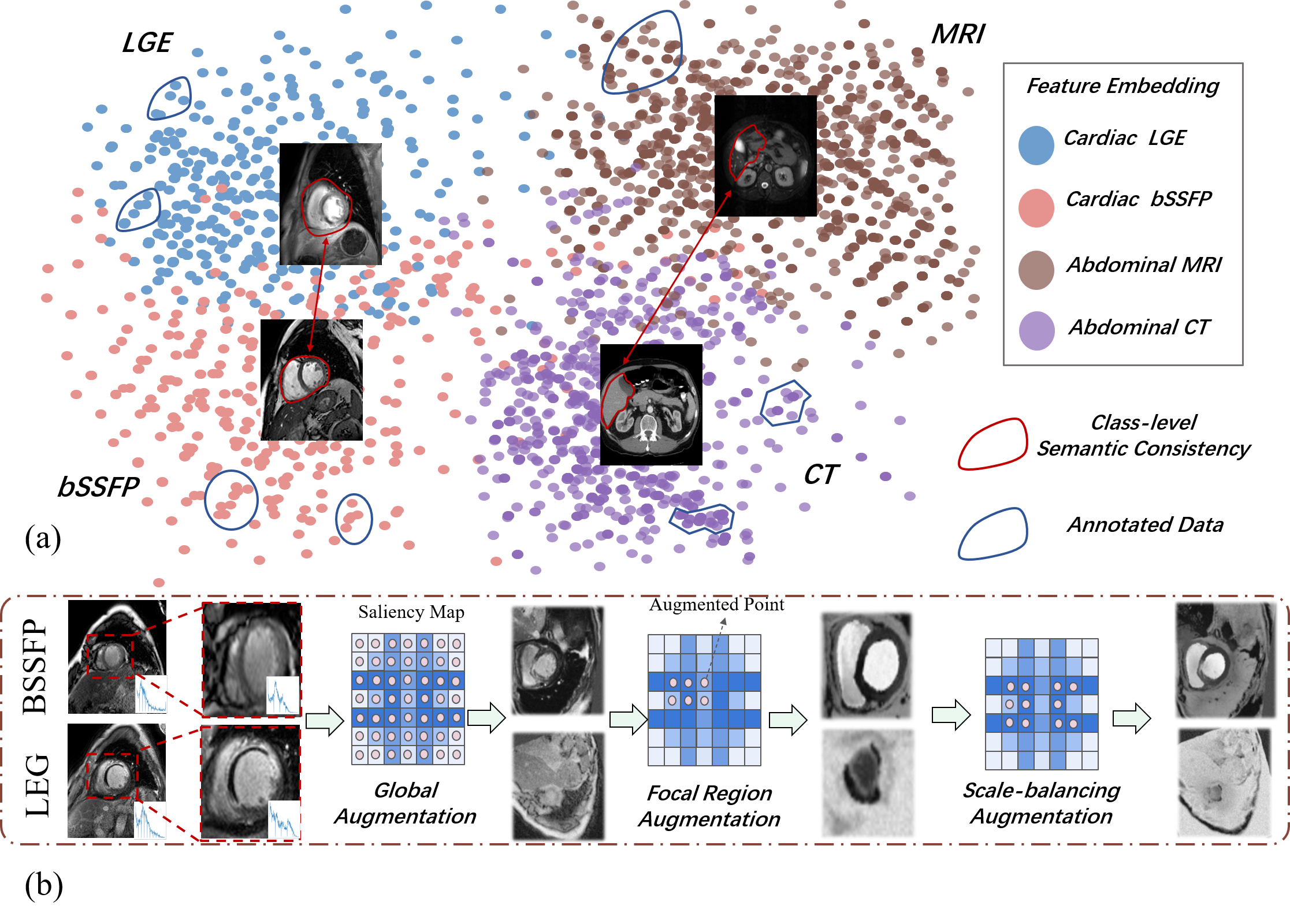} % Reduce the figure size so that it is slightly narrower than the column.
\caption{(a) Visualization of dataset features and labeled data, highlighting the semantic similarities of core regions across different modalities. (b) Overview of the proposed semantics-guided semi-supervised data augmentation.}
\label{fig2}
\end{figure}

Significant efforts have been dedicated to addressing the challenges posed by limited labeled data and domain shifts. SSL has been extensively investigated as a promising approach for training models by leveraging a small amount of labeled data alongside a much larger pool of unlabeled data. The success of SSL methods has facilitated their application in scenarios with limited annotations. However, SSL commonly assumes that labeled and unlabeled data are drawn from the same distribution, an assumption that often goes unverified in practice.  In practice, as illustrated in Fig. \ref{fig2}(a), labeled data represent only a small subset of the overall data distribution \cite{zhou2024simple}. While semantic information embedded in the limited labeled data is essential for accurate segmentation, this information frequently fails to generalize effectively due to domain bias. Furthermore, internal domain shifts and insufficient label-guided semantic information within a single dataset further impair performance. As shown in Fig. \ref{fig1}(b), even when the training and testing domains are identical, incorporating knowledge of domain shifts enhances SSL segmentation accuracy, underscoring the detrimental effects of internal domain shifts on model performance. Additionally, as demonstrated in Fig. \ref{fig1}(a) and (b), integrating domain shift knowledge leads to significant performance improvements across datasets with different modalities. Consequently, improving the generalization capability of existing SSL methods remains a paramount research direction.

A straightforward approach to mitigating domain shifts involves acquiring and labeling large and diverse datasets. However, this strategy is both costly and challenging, as is widely acknowledged in the community. Recent advancements, such as domain adaptation (DA) and domain generalization (DG), present promising alternatives. DA trains a model on source domains and enables generalization to a target domain when target domain data is available. In contrast, DG leverages information solely from multiple source domains. A stricter variant, SDG, relies exclusively on single-source domain data, making it particularly suitable to scenarios with limited access to medical images compared to DA and DG. However, these methods often assume fully labeled source domain data and struggle to maintain performance under label scarcity. The high costs associated with pixel-level labeling also limit its practical application, emphasizing the need for robust SDG methods that can function effectively with limited labeled data.

As discussed, prior efforts in SSL, DG, and SDG have primarily focused on isolated solutions. However, the simultaneous presence of limited annotations and domain shifts is a frequent challenge in clinical data. Developing deep learning models that can generalize to unseen domain data while effectively utilizing limited annotations offers significant practical value \cite{zhang2024s,ma2024constructing,yan2024prompt}. In this study, we address the challenging problem of semi-supervised single-domain generalization for medical image segmentation. Intuitively, directly integrating SSL and DG methods appears to be a promising solution. Existing DG and SDG methods, typically designed for fully supervised settings with abundant labeled data, primarily aim to expand the source data distribution to simulate domain shifts on the target domain. In SSL settings, where only limited labeled data is available, data augmentation expands the data distribution but often introduces noise. Consequently, SSL models with limited labeled data struggle to learn robust feature representations, resulting in domain-biased predictions and training instability as errors propagate during training.

To overcome the challenges of domain-biased learning and training instability with limited labeled data, we prioritize learning domain-invariant semantic features rather than simulating unseen domain distributions. We observe that explicit semantic knowledge for medical image segmentation exists in labeled data, and that the same anatomical regions exhibit class-level semantic consistency across domains. For example, as shown in Fig. \ref{fig2}, cardiac organs display strong semantic similarities across late gadolinium enhancement (LGE) and balanced steady-state free precession (bSSFP) imaging modalities. Domain shifts primarily arise from variations in pixel brightness mapping. These shifts can be modeled as linear combinations of $C$ (class number) random variables \cite{su2023rethinking,wang2024mln}. Additionally, non-interest regions provide explicit information about domain distributions. Extracting invariant semantic information while accounting for explicit distributional interference is crucial for improving domain generalization ability. Inspired by this observation, we apply distinct data augmentation strategies for global images and focal regions (as shown in Fig. \ref{fig2}(b)). Global augmentation (GA) enriches explicit domain distribution information, while focal region augmentation (FA) simulates class-specific shifts to promote domain-invariant feature learning. This approach transforms limited annotations into extensive domain knowledge while preserving domain-invariant semantic information. To ensure augmented images align with real-world distributions, we introduce an augmentation scale-balancing mechanism (SBA) that uses saliency feature maps during training to correct augmented data. Unlike existing data augmentation-based DG methods that focus exclusively on global images, our approach  expands the data distribution while preserving core regional semantic features, promoting domain-invariant feature learning under various perturbations.

To further improve domain-invariant feature learning, we employ a multi-decoder U-Net pipeline SSL network. Specifically, we introduce model perturbations across different decoders and incorporate a consistency loss based on uncertainty estimation. Uncertainty estimation from multiple decoders is integrated into the saliency map calculation. The ensemble predictions from the multi-decoder model are used as pseudo-labels to guide FA for unlabeled data. In this way, FA and SBA are extended to unlabeled data. Additionally, we adopt a deep mutual learning strategy to improve weight sharing across decoders, effectively mitigating oscillations during training.

The main contributions of this paper are summarized as follows: (\textbf{1}) We propose FSDA-DG, a novel framework that addresses the coexistence of domain shifts and label scarcity, thereby enhancing the model's applicability and robustness in real-world deployment scenarios. (\textbf{2}) We introduce a semantics-guided semi-supervised data augmentation approach that transforms both labeled and unlabeled data into extensive domain knowledge while preserving domain-invariant semantic information. (\textbf{3}) We design a multi-decoder U-Net pipeline SSL network featuring a consistency loss and a deep mutual learning strategy to facilitate domain-invariant feature learning. (\textbf{4}) Extensive experiments demonstrate that FSDA-DG achieves superior out-of-distribution generalization performance with limited source labeled data.

\section{Related work}
\subsection{Domain Generalization}
DG leverages source domain data to train models that can generalize directly to unseen target domains \cite{zhou2022generalizable,9540778,segu2023batch}. SDG extends this concept by developing models that can handle unseen domain shifts using only single-source domain data, making it an exceptionally challenging task \cite{su2023rethinking}. Techniques such as image style transformation and data augmentation, which indirectly address potential shifts on target domains, have demonstrated effectiveness for both DG and SDG tasks \cite{wu2020generalization,su2023rethinking,seo2020learning}.

In medical imaging, \cite{ouyang2022causality} introduced a data augmentation method aimed at removing spurious correlations through causal intervention. \cite{wang2024mln} proposed source-similar and source-dissimilar data augmentation methods to enrich the domain distribution of the training data. \cite{zhou2022generalizable} developed image style transformation with dual normalization to address cross-modality generalization tasks. However, these approaches primarily focus on augmenting global images, which limits their ability to accurately predict or simulate target domain distributions \cite{olsson2021classmix, zhang2021objectaug}.

Moreover, most DG and SDG methods require large amounts of source domain data and high-quality annotations, limiting their applicability in real-world clinical settings where pixel-level labeling is prohibitively expensive. When only limited labeled data is available, indiscriminate data augmentation often provides minimal performance gains and can even lead to incorrect learning due to a lack of semantic guidance \cite{garcea2023data}. In this work, we focus on learning domain-invariant features rather than simulating unseen domain data distributions to address cross-domain generalization challenges. To achieve this, FSDA-DG employs distinct data augmentation strategies for both global images and label-guided local regions, transforming data into extensive domain knowledge while preserving domain-invariant semantic information.

\subsection{Semi-Supervised Learning}
Pixel-wise annotation for medical images is particularly challenging and expensive. SSL has demonstrated significant potential in addressing these issues. By utilizing large volumes of unlabeled data, SSL enhances model performance and reduces reliance on costly manual annotations \cite{ma2024constructing}.

Existing SSL methods can be broadly divided into two main categories: (1) Pseudo-label-based SSL and (2) Consistency-based SSL. In pseudo-labeling SSL \cite{lee2013pseudo}, the model is initially trained on labeled data, and subsequently generates pseudo-labels for unlabeled data. These pseudo-labels are then iteratively refined and used to update the model. However, this approach often requires the application of multiple thresholds to mitigate the impact of noisy or incorrect pseudo-labels, which can cause training instability. Consistency-based SSL \cite{tarvainen2017mean,jeong2019consistency} promotes stability by enforcing consistency in the predictions of unlabeled data under various perturbations, such as horizontal flips, contrast changes, and brightness adjustments.

\cite{wu2022mutual} introduced mutual consistency loss to train parallel networks, while \cite{luo2022semi} employed uncertainty-rectified pyramid consistency to refine predictions for unlabeled data using multi-scale outputs. While these methods have shown success in label-scarce settings, their performance often degrades in the presence of domain shifts. As previously discussed, SSL approaches typically assume that labeled and unlabeled data share the same underlying data distribution. However, this assumption is frequently violated in real-world scenarios due to inherent domain variations. From a theoretical perspective, \cite{litowards} recently demonstrated that learning class-invariant features with limited annotations can significantly improve generalization performance. Building on this insight, FSDA-DG introduces a novel semantics-guided data augmentation strategy tailored for medical imaging, which is coupled with a multi-decoder consistency learning framework. The proposed data augmentation method transforms limited labeled data into comprehensive domain knowledge while preserving domain-invariant semantic information. Additionally, by incorporating multi-mode consistency learning, FSDA-DG promotes domain-invariant feature learning under diverse data and model perturbations, thereby enhancing both robustness and cross-domain generalization capability.

\subsection{Semi-supervised Learning with Out-of-Distribution Generalization}
Recent studies have highlighted the importance of SSL methods designed for out-of-distribution generalization in medical image segmentation \cite{ma2024constructing, liu2021semi, li2021semantic, yao2022enhancing}(e.g., Semi-supervised Medical Domain Generalization, Semi-supervised Medical Unsupervised Domain Adaptation). These approaches aim to extend the applicability of SSL methods to unseen target domains, a critical requirement in clinical practice. In this section, we review key advancements to contextualize our contributions.

\cite{ma2024constructing} introduced a pioneering framework termed "mixed domain semi-supervised medical image segmentation," which combined SSL with unsupervised domain adaptation. Although effective, this method requires multi-source domain data and prior knowledge of the target domain, limiting its scalability in diverse clinical scenarios.  \cite{liu2021semi} were the first to integrate SSL with DG, using a meta-learning approach with disentanglement to improve generalization capability. Similarly,  \cite{li2021semantic} proposed an SSL method for semantic segmentation that models the joint image-label distribution through a generative framework. However, these techniques also depend on multi-source domain data and domain labels, both of which are often unavailable in real-world clinical settings. To address these limitations,  \cite{yao2022enhancing} proposed a Fourier transform-based data augmentation strategy combined with confidence-aware cross pseudo-supervision, effectively eliminating the need for domain labels. Nevertheless, this approach still relies on multi-source domain data, making it resource-intensive and challenging to implement at scale.

In contrast, FSDA-DG tackles the more challenging problem of single-domain generalization for medical image segmentation with limited annotations from a single source domain. By relying solely on single-source domain data and eliminating the need for domain labels, FSDA-DG offers a practical and scalable solution for clinical deployment. It effectively addresses both domain shifts and label scarcity, providing a more effective framework for real-world applications.

\begin{figure*}[t]
\centering
\includegraphics[width=0.9\textwidth]{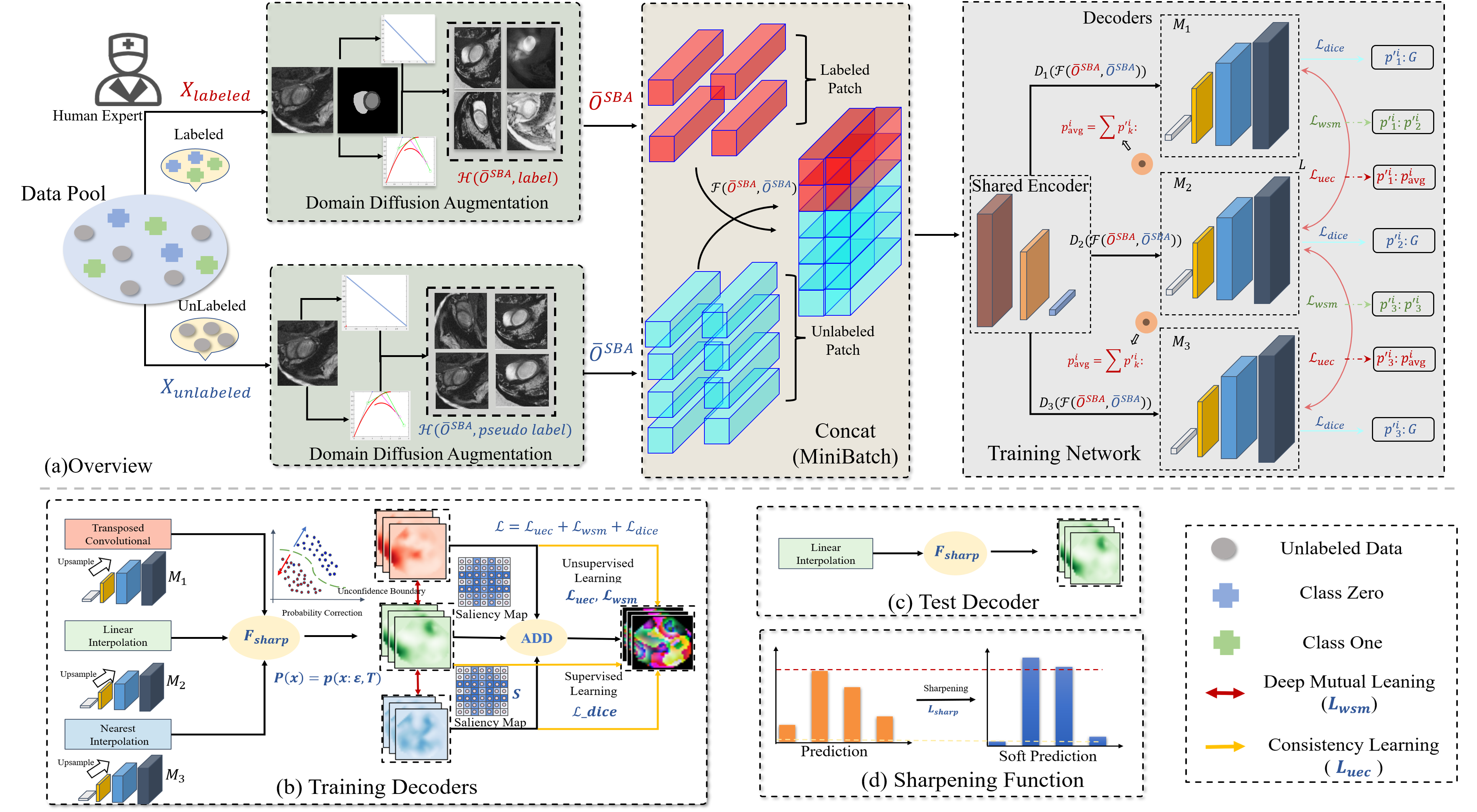} 
\caption{Diagram of our proposed FSDA-DG. FSDA-DG uses different data augmentation strategies for labeled and unlabeled data, then concats them within minibatch. During training, FSDA-DG uses a shared encoder and three independent decoders, whereas in the testing phase, only a single decoder is utilized.}
\label{fig3}
\vspace{-0.5cm}  
\end{figure*}

\section{Main Methodology}
To precisely define our approach, we first introduce key formulations. SSL uses both labeled data \((x_i, y_i) \in X_{\text{labeled}}\) and unlabeled data \(x_j \in X_{\text{unlabeled}}\) to improve the generalization and robustness of deep learning models. The training dataset is defined as \(X = X_{\text{labeled}} \cup X_{\text{unlabeled}}\). Given a small set of labeled data \(X_{\text{labeled}}\) and a large set of unlabeled data \(X_{\text{unlabeled}}\), the objective of SSL is to train a model \(M_s(\theta)\) that yields confident and accurate predictions, where \(\theta\) represents the parameters of model \(M_s\). SDG relies exclusively on source domain data \((x_i^s, y_i^s) \in X_{\text{labeled}}^s\) (where \(s\) denotes the source domain) to train a model \(M_d(\theta_s)\) that generalizes effectively to unseen target domains \(D^t\). FSDA-DG addresses a more challenging setting: it employs limited labeled source domain data \((x_i^s, y_i^s) \in X_{\text{labeled}}^s\) along with unlabeled source domain data \(x_s \in X_{\text{unlabeled}}^s\) (both from the same domain \(s\)) to train a model \(M_{sg}(\theta_{sg})\). The resulting model is designed to achieve superior generalization performance on unseen target domains \(D^t\).

This section outlines the proposed FSDA-DG framework. As illustrated in Fig.~\ref{fig3}, FSDA-DG comprises two main components:(1) a semantics-guided semi-supervised data augmentation strategy and (2) a multi-decoder U-Net pipeline network.

\begin{algorithm}[t]
\SetAlgoLined
\caption{Global Augmentation (GA)}
\KwIn{$x$: Input image}
\KwIn{$p_0$ and $p_3$: Starting and end points for B\'{e}zier curve transformation}
\KwOut{$O^{GA}(x)$: Global augmented output}

\BlankLine
\tcp{Generate a random number between 0 and 1}
$r \gets \text{Random()}$  

\eIf{$r > 0.1$}{
\tcp{Sample from Gaussian distribution }
    $\eta \sim \mathcal{T}\mathcal{N}(1, \sigma_1)$;
    
    $\mu \sim \mathcal{T}\mathcal{N}(0, \sigma_2)$;

    \tcp{Randomly generate two control points for the B\'{e}zier curve}
    $p_1 \gets \text{RandomPoint}()$\;
    $p_2 \gets \text{RandomPoint}()$\;
    \tcp{Apply B\'{e}zier curve transformation}
    $B(k) \gets \sum\nolimits_{i=0}^{I} P_i (1-w)^{I-i} w^i$\;
    $O^{GA}(x) \gets \eta  \Lambda(x) + \mu$\;
}{
    \tcp{Apply grayscale inversion transformation}
    $O^{GA}(x) \gets \eta  \Gamma(x) + \mu$\;
}
\Return $O^{GA}(x)$\;
\end{algorithm}

\subsection{Semantics-Guided Semi-Supervised Data Augmentation}
\subsubsection{Global and Focal Region Augmentation}
The coexistence of limited annotations and domain shifts is a prevalent challenge in clinical data. Deep learning models often overfit to specific data sources, which restricts their ability to generalize across modalities and impedes robust representation learning \cite{zhou2022generalizable}.Inspired by class-level representation learning and DG, we observe that explicit semantic knowledge exists in the labeled data, and that the same anatomical structures exhibit consistent class-level semantics across domains. Domain shifts primarily arise from variations in pixel brightness mappings, and features of an unseen target domain can be approximated as a linear combination of source domain features. Leveraging this insight, limited labeled data can be transformed into enriched domain knowledge, thereby improving generalization to unseen domains while reducing the reliance on extensive manual annotation. Additionally, non-target regions provide explicit information about the domain distribution. We therefore propose distinct augmentation strategies for target and non-target regions to promote the learning of domain-invariant features across diverse data distributions.

\begin{figure}[t]
\centering
\includegraphics[width=0.45\textwidth]{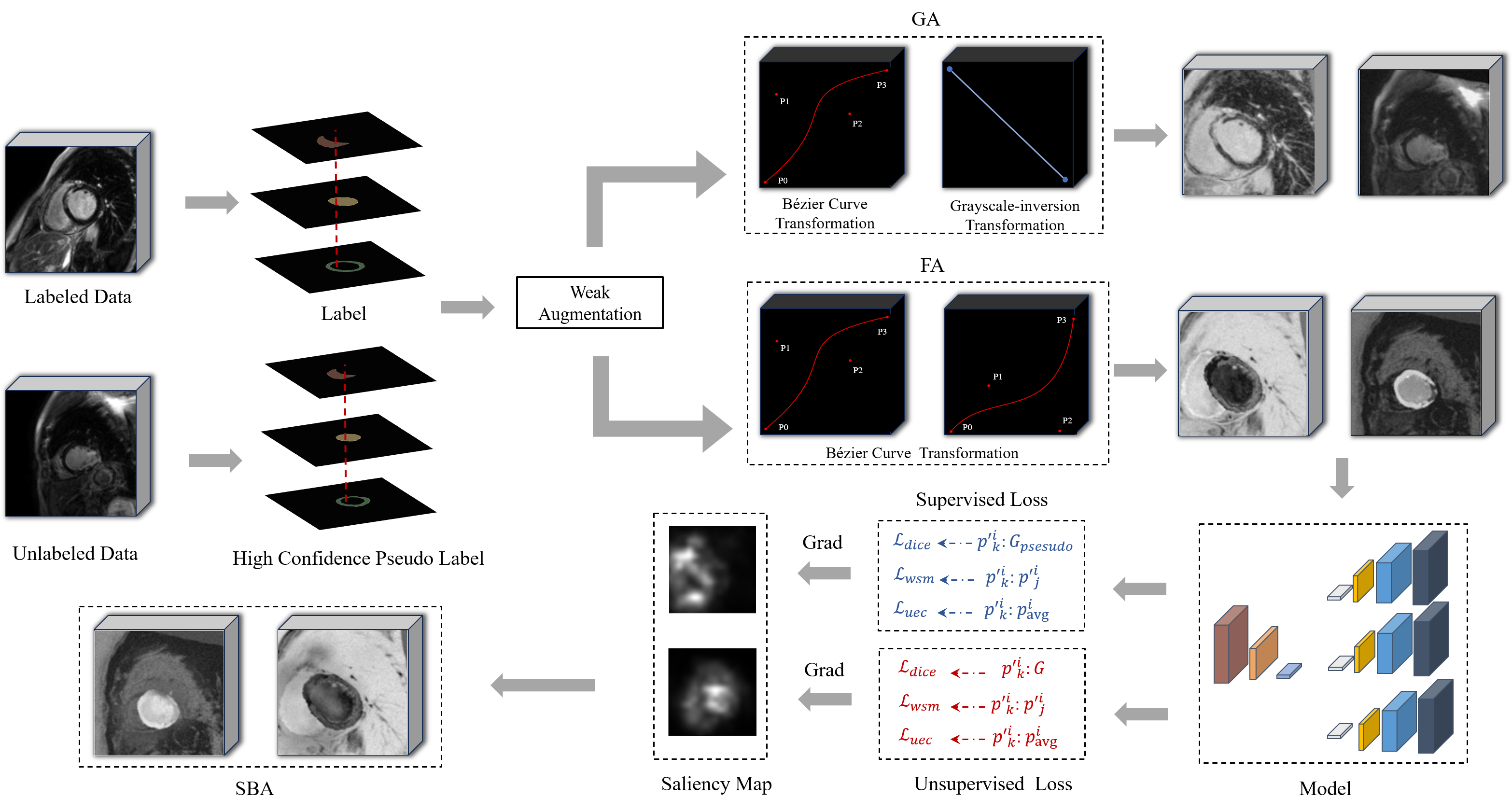} 
\caption{Overview of the proposed semantics-guided semi-supervised data augmentation, which applies distinct strategies to labeled and unlabeled data.}
\label{figFA_GA}
\end{figure}

Building on prior work \cite{su2023rethinking, zhou2022generalizable, wang2024mln}, we define a monotonic non-linear intensity transformation using the B\'{e}zier curves. The constrained B\'{e}zier curve is defined as:

\begin{equation}
\ B(k)=\sum\nolimits_{i=0}^{I}{{{P}_{i}}{{(1-w)}^{I-w}}{{w}^{i}}},I=3,w\in [0,1],\ 
\label{eq1}   
\end{equation}
where $w$ represents the ratio of the line's length, and $P_i$ denotes the control points, which are constrained within the range [0,1]. In this study, two sets of curve control points are randomly sampled from the range [0,1], with the starting and end points of both sets anchored at (0, 0) and (1, 1), respectively. Unlike prior work \cite{su2023rethinking}, which applies high-probability inversion transformations (swapping the start and end points to generate inverted images), FSDA-DG avoids such transformations in SSL settings, as they may introduce additional instability during model training. Instead, FSDA-DG adopts a simpler approach: a low-probability grayscale-inversion transformation to introduce controlled perturbations. This computationally efficient transformation reduces the model's sensitivity to brightness and contrast variations, encouraging it to focus on deeper semantic features. The grayscale-inversion transformation is defined as:

\begin{equation}
\ G_t=G_{max}-G_o,\
\label{eq3}   
\end{equation}
where $G_{max}$ represents the maximum pixel value of the image, and $G_o$ denotes the pixel value at position $o$. Using pixel-level transformations, we define Global Augmentation (GA) and Focal Region Augmentation (FA). Prior to augmentation, the pixel values of the input data are normalized to the range [0,1]. The monotonic non-linear transformation and grayscale-inversion transformation are denoted as $\Lambda\left(\cdot\right)$ and $\Gamma\left(\cdot\right)$, respectively. The GA is defined as:

\begin{equation}
O^{GA}(x) = \eta  \Lambda(x) + \mu \,
\label{eq4}
\end{equation}

\begin{equation}
O^{GA}(x) = \eta  \Gamma(x) + \mu \,
\label{eq5}
\end{equation}
where $\eta \sim \mathcal{T}\mathcal{N} \left(1, \sigma_1\right)$ and $\mu \sim \mathcal{T}\mathcal{N} \left(0, \sigma_2\right)$ are GA factors, with $\sigma_1$ and $\sigma_2$ representing the standard deviations of two truncated Gaussian distributions, respectively. $x$ denotes the input image. As shown in \textbf{Algorithm 1}, GA applies either a monotonic non-linear transformation or a grayscale-inversion transformation based on a specified probability. FA, leveraging the annotations of the training data, performs class-level augmentation, which is defined as:
\begin{equation}
\ O^{FA}=\sum_{n=0}^{N}\eta_n\psi_n\Lambda\left(x,m\right)+\mu_n, \ 
\label{eq6}
\end{equation}
where $\eta_n \sim \mathcal{T}\mathcal{N} \left(1,\sigma_1\right)$ and $\mu_n \sim \mathcal{T}\mathcal{N}\left(0,\sigma_2\right)$ are FA factors. $x$ denotes the input image, and $m$ represents the corresponding label. $N$ is the total number of classes, including the background class ($n=0$). To ensure that $O^{FA}$ differs from $O^{GA}$, we set $\psi_0 = 1$, while for $n \neq 0$, we set $\psi_n = 0.5$. In the FA setting, the background class encompasses all non-target anatomical regions within the image. Each class is augmented with independent parameters. Traditional mask-based data augmentation methods \cite{olsson2021classmix, zhang2021objectaug, su2023rethinking} rely on annotations for both focal regions and background, which are unavailable in SSL settings. We extend FA to unlabeled data by using pseudo-labels generated during training. \textbf{Algorithm 1} and \textbf{Algorithm 2} present the pseudo-code of the proposed GA and FA, respectively. The proposed augmentation effectively enriches the domain distribution under conditions of limited labeled data and mitigates domain bias.

\begin{algorithm}[t]
\SetAlgoLined
\caption{Focal Region Augmentation (FA)}
\KwIn{$x$: Input image}
\KwIn{$m$: Label or pseudo-label}
\KwIn{$p_0$ and $p_3$: Starting and end points for B\'{e}zier curve transformation}
\KwOut{$O^{FA}(x)$: Focal augmented output}

\BlankLine

\For{class $n = 0, \dots, N$}{
    \tcp{Set parameters based on whether $n$ is the background or a focal region class}
    $\psi_n \gets \begin{cases}
        1 & \text{if } n \text{ is background}, \\ 
        0.5 & \text{if } n \text{ is a focal region class}.
    \end{cases}$\;
    $\eta_n \sim \mathcal{T}\mathcal{N}(1, \sigma_1)$ \;
    $\mu_n \sim \mathcal{T}\mathcal{N}(0, \sigma_2)$ \;
    \tcp{Randomly generate two control points for the B\'{e}zier curve}
    $p_1 \gets \text{RandomPoint}()$\;
    $p_2 \gets \text{RandomPoint}()$\;
    $B(k) \gets \sum\nolimits_{i=0}^{I} P_i (1-w)^{I-i} w^i$\;
    \tcp{Apply the B\'{e}zier curve transformation for class $n$}
    $O^{FA}(x) \gets \eta_n \psi_n \Lambda(x, m) + \mu_n$\;
}

\Return $O^{FA}(x)$\;
\end{algorithm}

\subsubsection{Augmentation Scale-balancing Mechanism}
The random data augmentation of FA and GA introduces uncertainty due to a lack of constraints. This can cause augmented images to deviate from a realistic data distribution, which in turn may degrade model performance. To address this, we propose an augmentation scale-balancing mechanism that intelligently integrates the globally and focally augmented images. This mechanism combines gradient calculation with saliency feature mapping:

\begin{equation}
\mathcal{F} = \ \left({\bar{O}}^{GA}, {\bar{O}}^{FA}, \bar{m}\right) = \mathcal{H}\left(O^{GA}, O^{FA}, m\right), \
\label{eq7}   
\end{equation}

\begin{equation}
\ G = \ \nabla_{{\bar{O}}^{GA}} \mathcal{L}\left(M_\Phi\left({\bar{O}}^{GA}\right), \bar{m}\right),\
\label{eq8}   
\end{equation}

\begin{equation}
\ S = \Upsilon ({{L}_{2}}(G, g)), \
\label{eq9}   
\end{equation}

\begin{equation}
\ {\bar{O}}^{SBA} =\ S \cdot {\bar{O}}^{GA} + \left(1 - S\right) \cdot {\bar{O}}^{FA}. \
\label{eq10}   
\end{equation}
where $\mathcal{H}$ denotes the weak augmentation \cite{sohn2020fixmatch}, $m$ denotes annotations, $\mathcal{L}$ is the total training loss, and $M_\Phi$ is the segmentation model. $L_{2}$ refers to $l_{2}$ normalization, and $\Upsilon$ denotes a smoothing process that applies $l_{2}$ normalization to gradient values and downsamples the gradient map to a grid of size $g \times g$. The downsampled map is subsequently interpolated back to the original image size using quadratic B-spline kernels. A visual overview of this process is provided in Fig.~\ref{figFA_GA}. \textbf{Algorithm 3} provides the pseudo-code of the proposed SBA. The loss function for calculating the saliency map incorporates both supervised and unsupervised losses. The consistency loss, which is based on uncertainty estimation (described in Section 3.2), increases significantly when there is a large discrepancy between the predictions of different model decoders. Such an increase often indicates that the augmented data deviates from a realistic distribution. By incorporating this consistency loss into the saliency map calculation, FSDA-DG adaptively refines the augmentation strategy for unlabeled data.

SBA yields a final image that is a weighted linear combination of the globally and focally augmented images. Within this mechanism, the influence of FA is adaptively modulated by the saliency map, which is derived from model gradients. This balancing mechanism mitigates the training instability that can arise from using exclusively focal augmentation, particularly when such augmentations produce images that deviate significantly from a realistic data distribution. Therefore, we use ${\bar{O}}^{GA}$ and ${\bar{O}}^{SBA}$ as inputs.

\begin{algorithm}[t] % Use [t] for top placement
\SetAlgoLined
\caption{Scale-Balancing Mechanism (SBA)}
\KwIn{$\bar{x}$: Input image after weak augmentation}
\KwIn{$\bar{m}$: Label or pseudo-label}
\KwIn{$\bar{O}^{GA}(x)$, $\bar{O}^{FA}(x)$: GA and FA outputs after weak augmentation}
\KwIn{$g$: Grid size} 
\KwIn{$E$: Epoch threshold for using pseudo-labels}
\KwOut{$\bar{O}^{SBA}(x)$: SBA output}

\BlankLine

\eIf{$\bar{x} \in {{X}_{labeled}}$ \textbf{or} ($\bar{x} \in {{X}_{unlabeled}}$ \textbf{and} current epoch > $E$)}{
    \tcp{Compute gradient map}
    $G \gets \nabla_{\bar{O}^{GA}} \mathcal{L}\left(M_\Phi\left(\bar{O}^{GA}\right), \bar{m}\right)$ 
    
    \tcp{Compute saliency map}
    $s \gets \Upsilon\left({{L}_{2}}(G, g)\right)$
    
    \tcp{Interpolate saliency map to match image size}
    $S \gets \text{Interpolate}\left(s, \text{original size of } \bar{O}^{GA}\right)$
    
    \tcp{Combine global and focal augmentations}
    $\bar{O}^{SBA}(x) \gets S \cdot \bar{O}^{GA}(x) + (1 - S) \cdot \bar{O}^{FA}(x)$ 
}{
    $\bar{O}^{SBA}(x) \gets \bar{O}^{GA}(x)$
}

\BlankLine
\Return $\bar{O}^{SBA}(x)$\;

\end{algorithm}

\subsection{Consistency Loss Based on Uncertainty Estimation}
As shown in Fig. \ref{fig3}, FSDA-DG employs a multi-decoder architecture, where each decoder branch uses a distinct upsampling strategy. These distinct upsampling strategies function as model-level perturbations and provide two key advantages: (1) enhancing model robustness and (2) promoting the learning of domain-invariant features. Additionally, perturbations are introduced at both the data and model levels by applying distinct data augmentation and upsampling strategies. These dual perturbations encourage the learning of domain-invariant features that are consistent across the decoder branches, which is essential for addressing domain shifts in label-scarce scenarios. FSDA-DG also incorporates a shared encoder, which ensures that all decoders receive features with consistent high-level semantics, while also reducing computational costs.To enforce agreement between the decoders, we propose a consistency loss based on uncertainty estimation. Furthermore, we implement a deep mutual learning strategy to facilitate knowledge transfer between the decoders, thereby mitigating potential training instability.

\subsubsection{Sharpening Function}
Inspired by entropy minimization theory \cite{vu2019advent, grandvalet2004semi}, we introduce a sharpening function, $F_{sharp}$, to refine the model's output probability distribution.This function increases the model's predictive confidence by sharpening the probability distribution, effectively pushing predictions away from the decision boundary. Notably, $F_{sharp}$ operates as a post-processing step that adjusts output probabilities without incurring additional computational overhead. This approach simplifies the training pipeline and enhances the effectiveness of the proposed uncertainty-based consistency loss.

\begin{equation}
\small
 \resizebox{0.8\hsize}{!}{$
p'(y|x; \varepsilon_k) = \frac{p(y|x; \varepsilon_k)^{1/T}}
{p(y|x; \varepsilon_k)^{1/T} + \left (1 - p(y|x; \varepsilon_k)\right )^{1/T}}$},
\label{eq11}
\end{equation}
where $x$ is the input image, $\varepsilon_k$ represents the parameters of the network  $M_{k}$, and $p\left(y \middle| x; \varepsilon_k\right)$ denotes the network's predicted probability. The hyperparameter $T$ regulates the temperature of the sharpening function, where a lower value emphasizes confident predictions, while a higher value smooths the probability distribution.

\subsubsection{Uncertainty Estimation}
In FSDA-DG, we propose an uncertainty estimation method to evaluate the reliability of predictions across multiple branches. This method quantifies uncertainty by analyzing the variability in predicted probabilities across branches, allowing the identification of regions with inconsistent predictions. The resulting uncertainty information plays a crucial role in refining the model's learning process, particularly in semi-supervised learning (SSL) settings, where labeled data is limited.

To compute uncertainty, we first calculate the average predicted probability for pixel $i$ across $K$ branches:

\begin{equation}
\ p_{avg}^i=\frac{1}{K}\sum_{k=1}^{K}{p^\prime}_k^i, \
\label{eq12}
\end{equation}
where $K$ is the total number of segmentation branches, ${p^\prime}_k^i$ denotes the probability of pixel $i$ belonging to the target organ in branch $k$, and $p_{avg}^i$ represents the average probability of the outputs. Using this average prediction, the uncertainty for pixel $i$ in the branch $k$, denoted as $U_k^i$, is defined  as:

\begin{equation}
\ U_k^i={p^\prime}_k^i \cdot log\frac{{p^\prime}_k^i}{p_{avg}^i}, \
 \label{eq13}
\end{equation}
where $U_k^i$ quantifies the divergence between the sharpened probability ${p^\prime}_k^i$ and the average probability $p_{avg}^i$ to capture the disagreement among branches. The overall uncertainty for branch $k$, denoted as $U_k$, is subsequently computed by averaging the pixel-level uncertainties:

\begin{equation}
U_k = \frac{1}{I} \sum_{i=1}^{I} {U}_k^i, \
\label{eq13}
\end{equation}
where $I$ is the total number of pixels in the image. The uncertainty estimation process allows FSDA-DG to focus on regions of higher confidence, while accounting for areas with high variability, which are more likely to contain valuable learning signals. This enables the model to effectively balance learning from both labeled and unlabeled data, improving its overall performance on unseen domains. 

\subsubsection{Consistency Loss}
Based on the proposed uncertainty estimation, we define an uncertainty estimation loss $L_{une}$ and an uncertainty revised loss $L_{unr}$. The combination of $L_{une}$ and $L_{unr}$ constitutes the uncertainty estimation consistency loss, denoted as $L_{uec}$:

\begin{equation}
L_{uec} = \partial L_{une} + (1-\partial)L_{unr}, \
\label{eq14}
\end{equation}

\begin{equation}
L_{une} = \frac{1}{K} \sum_{k=1}^{K} U_k, \
\label{eq15}
\end{equation}

\begin{equation}
L_{unr} = \frac{1}{K} \sum_{k=1}^{K} \frac{\sum_{i=1}^{I} \| {p^\prime}_k^i - p_{avg}^i \|_2 \cdot \varpi_k^i}{\sum_{i=1}^{N} \varpi_k^i}, \
\label{eq16}
\end{equation}
where $\partial$ is a weighting parameter that balances the contributions of $L_{une}$ and $L_{unr}$, and $K$ denotes the total number of branches. $\varpi_k^i$ is a rectification weight, defined as $\varpi_k^i = e^{-U_k^i}$. $L_{unr}$ emphasizes reliable regions and ignores unreliable regions of predictions, while $L_{une}$ reduces prediction entropy. By combining these two terms, the total consistency loss, $L_{uec}$, improves the stability and effectiveness of the training process.

\begin{table*}[htbp]
  \centering
  \caption{Performance comparison of various methods on the target domain MRI and LGE. Dice score (\%) is used as the evaluation metric. $X_l$ represents the proportion of labeled data, and $X_u$ represents the proportion of unlabeled data. The best performance for 50\% labeled data is highlighted in bold, whereas the top performance for 20\% labeled data is underlined.}
  \begin{adjustbox}{width=1\textwidth}
    \begin{tabular}{lp{4.6em}lllllllllllllllll}
    \toprule
    \multicolumn{2}{l}{\multirow{2}[4]{*}{\textbf{Method}}} & \multicolumn{1}{l}{\textbf{Venue}} &   & \multicolumn{2}{l}{\textbf{\#Scans Used}} &   & \multicolumn{5}{l}{\textbf{Abdominal CT-MRI}} &   & \multicolumn{4}{l}{\textbf{Cardiac bSSFP-LGE}} \\
\cmidrule{3-3}\cmidrule{5-6}\cmidrule{8-12}\cmidrule{14-17}    \multicolumn{2}{l}{} &   &   & \multicolumn{1}{l}{\textbf{$X_l$}} & \multicolumn{1}{l}{\textbf{$X_u$}} &   & \multicolumn{1}{l}{\textbf{Liver}} & \multicolumn{1}{l}{\textbf{R-Kidney}} & \multicolumn{1}{l}{\textbf{L-Kidney}} & \multicolumn{1}{l}{\textbf{Spleen}} & \multicolumn{1}{l}{\textbf{Average}} &   & \multicolumn{1}{l}{\textbf{LVC}} & \multicolumn{1}{l}{\textbf{MYO}} & \multicolumn{1}{l}{\textbf{RVC}} & \multicolumn{1}{l}{\textbf{Average}} \\
\cmidrule{1-6}\cmidrule{8-12}\cmidrule{14-17}    \multicolumn{2}{l}{U-Net (target)} & \multicolumn{1}{l}{-} &   & 10\% & 0 &   & 69.90 & 72.65 & 70.42 & 70.76 & 70.93 &   & 58.03 & 50.83 & 62.77 & 57.21 \\
    \multicolumn{2}{l}{U-Net (target)} & \multicolumn{1}{l}{-} &   & 20\% & 0 &   & 83.51 & 83.73 & 79.86 & 82.39 & 82.37 &   & 75.45 & 74.02 & 71.45 & 73.64 \\
    \multicolumn{2}{l}{U-Net (target)} & \multicolumn{1}{l}{-} &   & 50\% & 0 &   & 86.19 & 88.09 & 83.12 & 86.95 & 86.10 &   & 84.83 & 78.55 & 78.64 & 80.67 \\
    \multicolumn{2}{l}{U-Net (target)} & \multicolumn{1}{l}{-} &   & 100\% & 0 &   & 91.30 & 92.43 & 89.86 & 89.83 & 90.85 &   & 92.04 & 83.11 & 89.30 & 88.15 \\
\cmidrule{1-6}\cmidrule{8-12}\cmidrule{14-17}    \multicolumn{1}{l}{} & Feddg & \multicolumn{1}{l}{CVPR’ 2021} &   & 20\% & 0 &   & 61.56 & 70.32 & 63.86 & 63.05 & 64.70 &   & 68.71 & 62.24 & 69.93 & 66.96 \\
    \multicolumn{1}{l}{} & Sadn & \multicolumn{1}{l}{CVPR’ 2022} &   &   &   &   & 64.51 & 64.85 & 69.92 & 62.33 & 65.40 &   & 72.63 & 64.44 & 71.79 & 69.62 \\
    \multicolumn{1}{l}{} & SLAug & \multicolumn{1}{l}{AAAI’ 2023} &   &   &   &   & 66.34 & 64.09 & 61.40 & 70.96 & 65.95 &   & 71.58 & 68.03 & 77.75 & 72.45 \\
    \multicolumn{1}{l}{} & LRSR & \multicolumn{1}{l}{TMI’ 2024} &   &   &   &   & 68.02 & 61.39 & 71.88 & 71.06 & 68.09 &   & 74.33 & 66.54 & 77.75 & 72.87 \\
    \multicolumn{1}{l}{D} & Feddg & \multicolumn{1}{l}{-} &   & 50\% & 0 &   & 73.98 & 78.71 & 78.98 & 75.03 & 76.68 &   & 81.73 & 77.42 & 77.94 & 79.03 \\
    \multicolumn{1}{l}{G} & Sadn & \multicolumn{1}{l}{-} &   &   &   &   & 83.45 & 82.91 & 80.41 & 73.01 & 79.95 &   & 85.37 & 74.62 & 82.78 & 80.92 \\
    \multicolumn{1}{l}{} & SLAug & \multicolumn{1}{l}{-} &   &   &   &   & 81.32 & 83.78 & 78.12 & 77.65 & 80.22 &   & \textbf{91.30} & 79.96 & 83.07 & 85.44 \\
    \multicolumn{1}{l}{} & LRSR & \multicolumn{1}{l}{-} &   &   &   &   & 83.62 & 83.99 & 81.69 & 74.32 & 80.91 &   & 90.67 & 81.34 & 81.01 & 84.35 \\
    \multicolumn{1}{l}{} & Feddg & \multicolumn{1}{l}{-} &   & 100\% & 0 &   & 77.66 & 77.35 & 81.81 & 80.37 & 79.30 &   & 84.71 & 78.95 & 82.46 & 82.04 \\
    \multicolumn{1}{l}{} & Sadn & \multicolumn{1}{l}{-} &   &   &   &   & 88.74 & 88.77 & 85.92 & 81.35 & 86.20 &   & 92.30 & 81.07 & 86.00 & 86.46 \\
    \multicolumn{1}{l}{} & SLAug & \multicolumn{1}{l}{-} &   &   &   &   & 90.08 & 89.23 & 87.54 & 87.67 & 88.63 &   & 91.53 & 80.65 & 87.90 & 86.69 \\
    \multicolumn{1}{l}{} & LRSR & \multicolumn{1}{l}{-} &   &   &   &   & 89.92 & 87.37 & 88.60 & 85.65 & 87.89 &   & 91.31 & 82.00 & 86.87 & 86.01 \\
\cmidrule{1-6}\cmidrule{8-12}\cmidrule{14-17}    \multicolumn{1}{l}{} & SASSNeT & \multicolumn{1}{l}{MICCAI’ 2020} &   & 10\% & 90\% &   & 28.07 & 29.87 & 21.91 & 30.46 & 27.58 &   & 8.79 & 29.60 & 21.55 & 19.98 \\
    \multicolumn{1}{l}{} & DTC & \multicolumn{1}{l}{AAAI’ 2021} &   &   &   &   & 30.40 & 17.33 & 35.02 & 16.84 & 24.90 &   & 38.12 & 38.22 & 45.91 & 40.75 \\
    \multicolumn{1}{l}{} & MC-Net+ & \multicolumn{1}{l}{MedIA’ 2022} &   &   &   &   & 33.02 & 20.82 & 35.35 & 26.01 & 28.80 &   & 33.92 & 32.03 & 29.02 & 31.66 \\

    \multicolumn{1}{l}{} & MLRPL & \multicolumn{1}{l}{MedIA' 2024} &   &   &   &   & 35.94 & 33.01 & 40.18 & 30.11 & 34.82 &  & 40.57 & 36.60 & 50.81 & 42.67 \\
    \multicolumn{1}{l}{S} & SASSNeT & \multicolumn{1}{l}{-} &   & 20\% & 80\% &   & 30.65 & 31.67 & 35.74 & 35.32 & 33.35 &   & 33.05 & 39.98 & 48.08 & 40.57 \\
    \multicolumn{1}{l}{S} & DTC & \multicolumn{1}{l}{-} &   &   &   &   & 33.79 & 22.50 & 38.10 & 28.09 & 30.62 &   & 44.83 & 50.24 & 53.33 & 49.47 \\
    \multicolumn{1}{l}{L} & MC-Net+ & \multicolumn{1}{l}{-} &   &   &   &   & 42.53 & 35.70 & 31.98 & 36.85 & 36.77 &   & 43.97 & 48.30 & 49.12 & 47.13 \\
    \multicolumn{1}{l}{} & MLRPL & \multicolumn{1}{l}{-} &   &   &   &   & 45.93 & 40.47 & 45.67 & 36.47 & 42.14
    &  & 48.85 & 47.30 & 55.72 & 50.62 \\
    \multicolumn{1}{l}{} & SASSNeT & \multicolumn{1}{l}{-} &   & 50\% & 50\% &   & 40.42 & 40.48 & 39.87 & 37.06 & 39.46 &   & 35.32 & 41.28 & 49.78 & 42.19 \\
    \multicolumn{1}{l}{} & DTC & \multicolumn{1}{l}{-} &   &   &   &   & 41.88 & 45.95 & 42.56 & 34.46 & 41.21 &   & 46.27 & 53.33 & 56.80 & 52.13 \\
    \multicolumn{1}{l}{} & MC-Net+ & \multicolumn{1}{l}{-} &   &   &   &   & 54.67 & 50.24 & 40.06 & 41.55 & 46.63 &   & 51.02 & 53.49 & 50.82 & 51.78 \\ 
   \multicolumn{1}{l}{} & MLRPL & \multicolumn{1}{l}{-} &   &   &   &   & 51.00 & 51.39 & 50.95 & 43.04 & 49.10
    &  & 54.47 & 57.61 & 57.97 & 56.68 \\
    \cmidrule{1-6}\cmidrule{8-12}\cmidrule{14-17}
     \multicolumn{1}{l}{S} & EPL & \multicolumn{1}{l}{AAAI’ 2022} &   & 10\% & 90\% &   & 70.81 & 72.33  & 68.45 & 72.05 & 70.91&   & 73.49 & 70.91 & 69.25 & 71.22 \\
     \multicolumn{1}{l}{S} & MIDSS & \multicolumn{1}{l}{CVPR’ 2024} &   &   &   &   & 59.55 & 53.79 & 68.64 & 59.98 & 60.49 &   & 64.61 & 60.80 & 68.75 & 64.72 \\
     \multicolumn{1}{l}{L} & EPL & \multicolumn{1}{l}{-} &   & 20\% & 80\% &   & 72.53 & 75.69 & 73.42 & 73.08 & 73.68&   & 75.82 & 71.90 & 73.85 & 73.86 \\
     \multicolumn{1}{l}{O} & MIDSS & \multicolumn{1}{l}{-} &   & &   &   & 69.75 & 69.39 & 73.88 & 70.06 & 70.77 &   & 72.74 & 72.50 & 75.36 & 73.53 \\
     \multicolumn{1}{l}{D} & EPL & \multicolumn{1}{l}{-} &   & 50\% & 50\% &   & 74.35 & 79.35  & 75.00 & 75.76 & 76.12&   & 81.47& 81.66 & 80.93 & 81.35 \\
     \multicolumn{1}{l}{G} & MIDSS & \multicolumn{1}{l}{-} &   &   &   &   & 74.99 & 79.46 & 82.50 &  \textbf{85.00} & 80.49 & & 80.37 & \textbf{82.55} & 84.87 & 82.60 \\

\cmidrule{1-6}\cmidrule{8-12}\cmidrule{14-17}    \multicolumn{1}{l}{O} & FSDA-DG &   &   & 10\% & 90\% &   & 79.36 & 82.16 & 79.18 & 80.70 & 80.35 &   & 86.24 & 73.25 & 76.76 & 78.75 \\
    \multicolumn{1}{l}{U} & FSDA-DG &   &   & 20\% & 80\% &   & \underline{85.56} & \underline{85.50} & \underline{81.98} & \underline{82.82} & \underline{83.97} &   & \underline{90.41} & \underline{78.84} & \underline{84.41} & \underline{84.56} \\
    \multicolumn{1}{l}{R} & FSDA-DG &   &   & 50\% & 50\% &   & \textbf{87.59} & \textbf{85.75} & \textbf{85.69} & 84.93 & \textbf{85.99} &   & 91.18 & 80.11 & \textbf{86.33} & \textbf{85.87} \\
    \bottomrule
    \end{tabular}%
\end{adjustbox}
\label{Tab1}%
\end{table*}%

\begin{table*}[htbp]
  \centering
  \caption{Performance comparison of various methods on the target domain CT and bSSFP. Dice score (\%) is used as the evaluation metric. $X_l$ represents the proportion of labeled data, and $X_u$ represents the proportion of unlabeled data. The best performance for 50\% labeled data is highlighted in bold, whereas the top performance for 20\% labeled data is underlined.}
  \begin{adjustbox}{width=1\textwidth}
    \begin{tabular}{lp{5em}lllllllllllllllll}
    \toprule
    \multicolumn{2}{l}{\multirow{2}{*}{\textbf{Method}}} & \multicolumn{1}{l}{\textbf{Venue}} &   & \multicolumn{2}{l}{\textbf{\#Scans Used}} &   & \multicolumn{5}{l}{\textbf{Abdominal MRI-CT}} &   & \multicolumn{4}{l}{\textbf{Cardiac LGE-bSSFP}} \\
\cmidrule{3-3}\cmidrule{5-6}\cmidrule{8-12}\cmidrule{14-17}    \multicolumn{2}{l}{} &   &   & \multicolumn{1}{l}{\textbf{$X_l$}} & \multicolumn{1}{l}{\textbf{$X_u$}} &   & \multicolumn{1}{l}{\textbf{Liver}} & \multicolumn{1}{l}{\textbf{R-Kidney}} & \multicolumn{1}{l}{\textbf{L-Kidney}} & \multicolumn{1}{l}{\textbf{Spleen}} & \multicolumn{1}{l}{\textbf{Average}} &   & \multicolumn{1}{l}{\textbf{LVC}} & \multicolumn{1}{l}{\textbf{MYO}} & \multicolumn{1}{l}{\textbf{RVC}} & \multicolumn{1}{l}{\textbf{Average}} \\
\cmidrule{1-6}\cmidrule{8-12}\cmidrule{14-17}
    \multicolumn{2}{l}{U-Net (target)} & \multicolumn{1}{l}{-} &   & 10\% & 0 &   & 69.50 & 73.72 & 76.61 & 70.94 & 72.69 &   & 70.85 & 58.70 & 60.05 & 63.20 \\
    \multicolumn{2}{l}{U-Net (target)} & \multicolumn{1}{l}{-} &   & 20\% & 0 &   & 80.32 & 84.05 & 86.24 & 80.32 & 82.73 &   & 79.62 & 72.66 & 79.42 & 77.25 \\
    \multicolumn{2}{l}{U-Net (target)} & \multicolumn{1}{l}{-} &   & 50\% & 0 &   & 85.14 & 88.94 & 88.01 & 85.36 & 86.86 &   & 85.27 & 80.75 & 85.90 & 83.97 \\
    \multicolumn{2}{l}{U-Net (target)} & \multicolumn{1}{l}{-} &   & 100\% & 0 &   & 98.87 & 92.11 & 91.75 & 88.55 & 89.74 &   & 91.16 & 82.93 & 90.39 & 88.16 \\
\cmidrule{1-6}\cmidrule{8-12}\cmidrule{14-17}
    \multicolumn{1}{l}{} & Feddg & \multicolumn{1}{l}{CVPR’ 2021} &   & 20\% & 0 &   & 51.86 & 49.67 & 52.92 & 61.72 & 54.04 &   & 50.74 & 40.91 & 52.86 & 48.17 \\
    \multicolumn{1}{l}{} & Sadn & \multicolumn{1}{l}{CVPR’ 2022} &   &   &   &   & 67.51 & 62.80 & 71.84 & 75.01 & 69.29 &   & 62.33 & 54.32 & 67.30 & 61.32 \\
    \multicolumn{1}{l}{} & SLAug & \multicolumn{1}{l}{AAAI’ 2023} &   &   &   &   & 68.75 & 62.13 & 74.64 & \underline{77.63} & 70.79 &   & 58.77 & 59.53 & 67.33 & 61.88 \\
    \multicolumn{1}{l}{} & LRSR & \multicolumn{1}{l}{TMI’ 2024} &   &   &   &   & 71.88 & 63.34  &72.06 & 73.85 & 70.28 &   & 65.90 & 66.54 & 70.30 & 67.58 \\
    \multicolumn{1}{l}{D} & Feddg & \multicolumn{1}{l}{-} &   & 50\% & 0 &   & 70.39 & 71.02 & 72.74 & 77.89 & 73.01 &   & 61.52 & 63.69 & 70.33 & 65.18 \\
    \multicolumn{1}{l}{G} & Sadn & \multicolumn{1}{l}{-} &   &   &   &   & 77.46 & \textbf{86.56} & 84.00 & 80.78 & 82.20 &   & 71.36 & 60.24 & 70.06 & 67.22 \\
    \multicolumn{1}{l}{} & SLAug & \multicolumn{1}{l}{-} &   &   &   &   & 82.87 & 84.92 & 81.33 & 80.90 & 82.51 &   & 74.86 & 70.02 & 78.54 & 74.47 \\
    \multicolumn{1}{l}{} & LRSR & \multicolumn{1}{l}{-} &   &   &   &   & 81.31 & 81.98 & 80.29 &  83.66 & 81.81 &   & 75.25 & 71.97 & 80.46 & 75.89 \\
    \multicolumn{1}{l}{} & Feddg & \multicolumn{1}{l}{-} &   & 100\% & 0 &   & 83.65 & 82.92 & 79.89 & 82.07 & 82.13 &   & 75.69 & 73.02 & 75.48 & 74.73 \\
    \multicolumn{1}{l}{} & Sadn & \multicolumn{1}{l}{-} &   &   &   &   & 89.79 & 88.02 & 83.10 & 86.41 & 86.83 &   & 85.39 & 81.61 & 86.80 & 84.60 \\
    \multicolumn{1}{l}{} & SLAug & \multicolumn{1}{l}{-} &   &   &   &   & 90.08 & 86.23 & 84.54 & 85.67 & 86.63 &   & 91.53 & 80.65 & 87.90 & 86.69 \\
    \multicolumn{1}{l}{} & LRSR & \multicolumn{1}{l}{-} &   &   &   &    & 86.36 & 89.61 & 84.94 & 87.91 &    87.21  &   & 86.45 &79.99 & 88.68 & 85.04 \\
\cmidrule{1-6}\cmidrule{8-12}\cmidrule{14-17}
    \multicolumn{1}{l}{} & SASSNeT & \multicolumn{1}{l}{MICCAI’ 2020} &   & 10\% & 90\% &   & 15.89 & 16.69 & 27.35 & 22.01 & 20.49 &   & 11.26 & 28.79 & 27.35 & 22.53 \\
    \multicolumn{1}{l}{} & DTC & \multicolumn{1}{l}{AAAI’ 2021} &   &   &   &   & 19.86 & 15.01 & 29.83 & 19.77 & 21.12 &   & 20.58 & 35.98 & 37.53 & 31.36 \\
    \multicolumn{1}{l}{} & MC-Net+ & \multicolumn{1}{l}{MedIA’ 2022} &   &   &   &   & 20.47 & 20.62 & 27.05 & 18.89 & 21.76 &   & 34.84 & 35.54 & 29.72 & 35.97 \\
    \multicolumn{1}{l}{} & MLRPL & \multicolumn{1}{l}{MedIA'2024} &   &   &   &   & 21.64 & 25.85 & 33.05 & 23.70 & 26.07 &   & 41.05 & 31.28 & 33.26 & 35.20 \\
    \multicolumn{1}{l}{S} & SASSNeT & \multicolumn{1}{l}{-} &   & 20\% & 80\% &   & 25.74 & 23.88 & 33.04 & 28.92 & 27.90 &   & 28.75 & 39.02 & 42.11 & 36.63 \\
    \multicolumn{1}{l}{S} & DTC & \multicolumn{1}{l}{-} &   &   &   &   & 30.02 & 39.91 & 30.70 & 30.76 & 32.85 &   & 38.00 & 42.08 & 49.68 & 43.25 \\
    \multicolumn{1}{l}{L} & MC-Net+ & \multicolumn{1}{l}{-} &   &   &   &   & 40.32 & 38.01 & 43.99 & 38.65 & 40.24 &   & 42.57 & 42.38 & 48.70 & 44.55 \\
    \multicolumn{1}{l}{} & MLRPL & \multicolumn{1}{l}{MedIA'2024} &   &   &   &   & 37.09 & 35.29 & 47.60 & 37.74 & 40.44 &   & 48.66 & 42.86 & 45.30 & 45.61 \\
    \multicolumn{1}{l}{} & SASSNeT & \multicolumn{1}{l}{-} &   & 50\% & 50\% &   & 28.45 & 30.95 & 37.47 & 37.04 & 33.48 &   & 31.91 & 40.67 & 44.78 & 39.12 \\
    \multicolumn{1}{l}{} & DTC & \multicolumn{1}{l}{-} &   &   &   &   & 35.94 & 50.94 & 33.66 & 32.09 & 38.16 &   & 41.42 & 43.61 & 51.23 & 45.42 \\
    \multicolumn{1}{l}{} & MC-Net+ & \multicolumn{1}{l}{-} &   &   &   &   & 54.67 & 53.24 & 45.08 & 42.30 & 48.82 &   & 43.56 & 52.96 & 50.94 & 49.15 \\ 
    \multicolumn{1}{l}{} & MLRPL & \multicolumn{1}{l}{MedIA'2024} &   &   &   &   & 49.90 & 54.06 & 51.15 & 43.48 & 49.90 &   & 54.00 & 50.79 & 52.01 & 52.27 \\
\cmidrule{1-6}\cmidrule{8-12}\cmidrule{14-17}
    \multicolumn{1}{l}{S} & EPL & \multicolumn{1}{l}{AAAI’ 2022} &   & 10\% & 90\% &   & 71.42 & 61.92 & 59.37 & 64.93 & 64.64 &   & 64.85 & 62.30 & 69.27 & 65.47 \\
    \multicolumn{1}{l}{S} & MIDSS & \multicolumn{1}{l}{CVPR’ 2024} &   &   &   &   & 53.99 & 53.45 & 56.04 & 67.77 & 57.81 &   & 60.16 & 57.44 & 59.75 & 59.12 \\
    \multicolumn{1}{l}{L} & EPL & \multicolumn{1}{l}{-} &   & 20\% & 80\% &   & 74.09 & 71.92 & 67.88 & 71.06 & 71.24 &   & 70.48 & 73.69 & 77.20 & 73.79 \\
    \multicolumn{1}{l}{O} & MIDSS & \multicolumn{1}{l}{-} &   &   &   &   & 69.20 & 65.41 & 64.94 & 76.77 & 69.08 &   & 59.33 & 69.80 & 71.29 & 66.80 \\
    \multicolumn{1}{l}{D} & EPL & \multicolumn{1}{l}{-} &   & 50\% & 50\% &   & 75.62 & 74.06 & 75.64 & 71.06 & 74.08 &   & 77.32 & 76.66 & 80.46 & 75.81 \\
    \multicolumn{1}{l}{G} & MIDSS & \multicolumn{1}{l}{-} &   &   &   &   & 78.41 & 78.91 & 80.94 & \textbf{89.13} & 81.82 &   & 70.49 & 77.94 & 82.37 & 76.93 \\
\cmidrule{1-6}\cmidrule{8-12}\cmidrule{14-17}
    \multicolumn{1}{l}{O} & FSDA-DG &   &   & 10\% & 90\% &   & 74.59 & 62.98 & 65.54 & 64.45 & 66.89 &   & 80.53 & 73.79 & 78.02 & 77.45 \\
    \multicolumn{1}{l}{U} & FSDA-DG &   &   & 20\% & 80\% &   & \underline{80.52} & \underline{78.89} & \underline{77.26} & 71.25 & \underline{76.98} &   & \underline{88.75} & \underline{77.11} & \underline{83.59} & \underline{83.15} \\
    \multicolumn{1}{l}{R} & FSDA-DG &   &   & 50\% & 50\% &   & \textbf{88.28} & 80.41 & \textbf{85.92} & 85.27 & \textbf{84.97} &   & \textbf{90.92} & \textbf{80.02} & \textbf{85.05} & \textbf{85.33} \\
    \bottomrule
    \end{tabular}%
  \end{adjustbox}
  \label{Tab2}%
\end{table*}%

% add data
\begin{table*}[htbp]
  \centering
  \caption{Performance comparison of various methods on the target domain MRI and LGE. Dice score (\%), IoU (\%), and HD95 (mm) are used as the evaluation metrics. $\pm$ denotes the standard deviation. \(^{*}\) denotes statistical significance (\(p < 0.05\)) between the Dice score, IoU, or HD95 of a given method and that of our method, as determined by a paired t-test. $X_l$ represents the proportion of labeled data, and $X_u$ represents the proportion of unlabeled data.}
  \begin{adjustbox}{width=1\textwidth}
    \begin{tabular}{lp{4.6em}lllllllllllllll}
    \toprule
    \multicolumn{2}{l}{\multirow{2}[4]{*}{\textbf{Method}}} &   & \multicolumn{2}{l}{\textbf{\#Scans Used}} &   & \multicolumn{3}{l}{\textbf{Abdominal CT-MRI}} &   & \multicolumn{3}{l}{\textbf{Cardiac bSSFP-LGE}} \\
    \cmidrule{4-5}\cmidrule{7-9}\cmidrule{11-13}    
    \multicolumn{2}{l}{} &   & \multicolumn{1}{l}{\textbf{$X_l$}} & \multicolumn{1}{l}{\textbf{$X_u$}} &   & \multicolumn{1}{l}{\textbf{Dice (\%)}} & \multicolumn{1}{l}{\textbf{IoU (\%)}} & \multicolumn{1}{l}{\textbf{HD95 (mm)}} &   & \multicolumn{1}{l}{\textbf{Dice (\%)}} & \multicolumn{1}{l}{\textbf{IoU (\%)}} & \multicolumn{1}{l}{\textbf{HD95 (mm)}} \\
    \cmidrule{1-5}\cmidrule{7-9}\cmidrule{11-13}
    \multicolumn{2}{l}{Feddg} &   
       & 10\% 
       & 0 &   
       & $52.91 \pm 15.81^{*}$
       & $43.77 \pm 12.79^{*}$
       & $25.11 \pm 20.16^{*}$
       &   
       & $58.47 \pm 10.37^{*}$
       & $50.04 \pm 12.60^{*}$
       & $28.17 \pm 28.48^{*}$ \\     
    \multicolumn{2}{l}{Sadn} &
    &   &   &   
    & $52.49 \pm 18.30^{*}$
    & $44.17 \pm 15.62^{*}$
    & $24.94 \pm 22.64^{*}$
    &   
    & $59.89 \pm 9.20^{*}$
    & $51.06 \pm 10.52^{*}$
    & $20.06 \pm 24.19^{*}$\\
    \multicolumn{2}{l}{SLAug} &   
    &   &   &   
    & $60.01 \pm 10.15^{*}$
    & $48.93 \pm 12.98^{*}$
    & $17.67 \pm 18.22^{*}$
    &   
    & $64.19 \pm 7.50^{*}$
    & $51.67 \pm 7.07^{*}$
    & $14.21 \pm 15.04^{*}$ \\
    \multicolumn{2}{l}{LRSR} &   
    &   &   &
    & $59.11 \pm 11.52^{*}$
    & $47.41 \pm 14.28^{*}$
    & $17.24 \pm 21.00^{*}$
    &   
    & $65.19 \pm 9.25^{*}$
    & $51.25 \pm 15.00^{*}$
    & $14.06 \pm 9.54^{*}$\\
    \multicolumn{2}{l}{EPL}&  
    & 10\%  & 90\%  &   
    & $70.91 \pm 16.17^{*}$
    & $60.66 \pm 17.19^{*}$
    & $16.36 \pm 21.34^{*}$
    &   
    & $71.22 \pm 12.38^{*}$
    & $60.60 \pm 9.57^{*}$
    & $12.88 \pm 17.20^{*}$ \\  
    \multicolumn{2}{l}{MIDSS}&  
    &   &   &   
    & $60.49 \pm 8.37^{*}$
    & $52.86 \pm 13.60^{*}$
    & $15.87 \pm 11.41^{*}$
    &   
    & $64.72 \pm 8.25^{*}$
    & $54.00 \pm 14.15^{*}$
    & $14.08 \pm 9.19^{*}$ \\
    \multicolumn{2}{l}{FSDA-DG}&  
    &   &   &   
    & $80.35 \pm 13.20$
    & $67.02 \pm 15.18$
    & $10.26 \pm 9.05$
    &   
    & $78.75 \pm 8.18$
    & $67.92 \pm 10.06$
    & $10.12 \pm 12.53$ \\    
\cmidrule{1-5}\cmidrule{7-9}\cmidrule{11-13}
     \multicolumn{2}{l}{Feddg} &  
     & 20\% & 0 &   
     & $64.70 \pm 8.36^{*}$
     & $55.26 \pm 10.04^{*}$
     & $25.51 \pm 20.21^{*}$
     &   
     & $66.96 \pm 6.56^{*}$
     & $54.87 \pm 13.06^{*}$
     & $23.40 \pm 17.45^{*}$ \\
    \multicolumn{2}{l}{Sadn} &
    &   &   &   
    & $65.40 \pm 15.34^{*}$
    & $52.36 \pm 9.89^{*}$
    & $20.00 \pm 17.75^{*}$
    &   
    & $69.62 \pm 11.82^{*}$
    & $61.98 \pm 10.57^{*}$
    & $16.74 \pm 23.24^{*}$\\
    \multicolumn{2}{l}{SLAug} &   
    &   &   &   
    & $65.95 \pm 7.54^{*}$
    & $55.96 \pm 14.38^{*}$
    & $16.24 \pm 10.31^{*}$
    &   
    & $72.45 \pm 9.50^{*}$
    & $61.05 \pm 10.39^{*}$
    & $14.11 \pm 10.00^{*}$ \\
    \multicolumn{2}{l}{LRSR} &   
    &   &   &
    & $68.09 \pm 9.34^{*}$
    & $59.39 \pm 13.01^{*}$
    & $14.14 \pm 12.58^{*}$
    &   
    & $72.87 \pm 9.25^{*}$
    & $61.25 \pm 15.00^{*}$
    & $14.06 \pm 9.54^{*}$\\
    \multicolumn{2}{l}{EPL}&  
    & 20\%  & 80\%  &   
    & $73.68 \pm 17.20^{*}$
    & $65.02 \pm 14.09^{*}$
    & $15.24 \pm 22.80^{*}$
    &   
    & $73.86 \pm 7.15^{*}$
    & $63.65 \pm 11.29^{*}$
    & $11.20 \pm 17.91^{*}$ \\
    \multicolumn{2}{l}{MIDSS}&  
    &   &   &   
    & $70.77 \pm 11.23^{*}$
    & $62.34 \pm 15.49^{*}$
    & $14.81 \pm 14.32^{*}$
    &   
    & $73.53 \pm 7.78^{*}$
    & $59.07 \pm 11.16^{*}$
    & $12.58 \pm 12.02^{*}$ \\
    \multicolumn{2}{l}{FSDA-DG}&  
    &   &   &   
    & $83.97 \pm 10.07$
    & $74.05 \pm 8.32$
    & $9.68 \pm 11.43$
    &   
    & $84.56 \pm 5.79$
    & $74.60 \pm 7.82$
    & $11.26 \pm 9.99$ \\
    \cmidrule{1-5}\cmidrule{7-9}\cmidrule{11-13}
    \multicolumn{2}{l}{Feddg} &  
     & 50\% & 0 &   
     & $76.68 \pm 9.25^{*}$
     & $66.65 \pm 12.08^{*}$
     & $11.91 \pm 12.36^{*}$
     &   
     & $79.03 \pm 8.26^{*}$
     & $65.41 \pm 10.45^{*}$
     & $17.38 \pm 17.28^{*}$ \\
    \multicolumn{2}{l}{Sadn} &
    &   &   &   
    & $79.95 \pm 14.76^{*}$
    & $68.49 \pm 10.25^{*}$
    & $16.98 \pm 15.34^{*}$
    &   
    & $80.92 \pm 10.01^{*}$
    & $71.23 \pm 10.05^{*}$
    & $14.45 \pm 15.77^{*}$\\
    \multicolumn{2}{l}{SLAug} &   
    &   &   &   
    & $80.22 \pm 7.52^{*}$
    & $66.90 \pm 12.85^{*}$
    & $13.06 \pm 9.40^{*}$
    &   
    & $85.44 \pm 10.22^{*}$
    & $75.31 \pm 9.07^{*}$
    & $11.64 \pm 8.55^{*}$ \\
    \multicolumn{2}{l}{LRSR} &   
    &   &   &
    & $80.91 \pm 8.28^{*}$
    & $65.47 \pm 11.61^{*}$
    & $11.45 \pm 13.00^{*}$
    &   
    & $84.35 \pm 10.61^{*}$
    & $73.11 \pm 13.62^{*}$
    & $14.39 \pm 10.54^{*}$\\   
    \multicolumn{2}{l}{EPL}&  
    & 50\%  & 50\%  &   
    & $76.12 \pm 15.81^{*}$
    & $67.89 \pm 16.35^{*}$
    & $12.94 \pm 16.46^{*}$
    &   
    & $81.35 \pm 6.60$
    & $73.25 \pm 9.68$
    & $15.73 \pm 14.04^{*}$ \\   
    \multicolumn{2}{l}{MIDSS}&  
    &   &   &   
    & $80.49 \pm 9.62^{*}$
    & $69.93 \pm 14.28^{*}$
    & $10.05 \pm 11.81$
    &   
    & $82.60 \pm 9.00^{*}$
    & $71.38 \pm 10.74^{*}$
    & $12.08 \pm 9.28^{*}$ \\
    \multicolumn{2}{l}{FSDA-DG}&  
    &   &   &   
    & $85.99 \pm 10.55$
    & $76.03 \pm 10.43$
    & $8.54 \pm 10.06$
    &   
    & $85.87 \pm 5.11$
    & $76.20 \pm 7.29$
    & $8.43 \pm 8.07$ \\
    \bottomrule
    \end{tabular}%
  \end{adjustbox}
  \label{Tab3}%
\end{table*}%

% add data
\begin{table*}[htbp]
  \centering
  \caption{Performance comparison of various methods on the target domain CT and bSSFP. Dice score (\%), IoU (\%), and HD95 (mm) are used as evaluation metrics. $\pm$ denotes the standard deviation. \(^{*}\) denotes statistical significance (\(p < 0.05\)) between the Dice score, IoU, or HD95 of a given method and that of our method, as determined by a paired t-test. $X_l$ represents the proportion of labeled data, and $X_u$ represents the proportion of unlabeled data.}
  \begin{adjustbox}{width=1\textwidth}
    \begin{tabular}{lp{4.6em}lllllllllllllll}
    \toprule
    \multicolumn{2}{l}{\multirow{2}[4]{*}{\textbf{Method}}} &   & \multicolumn{2}{l}{\textbf{\#Scans Used}} &   & \multicolumn{3}{l}{\textbf{Abdominal MRI-CT}} &   & \multicolumn{3}{l}{\textbf{Cardiac LGE-bSSFP}} \\
    \cmidrule{4-5}\cmidrule{7-9}\cmidrule{11-13}    
    \multicolumn{2}{l}{} &   & \multicolumn{1}{l}{\textbf{$X_l$}} & \multicolumn{1}{l}{$X_u$} &   & \multicolumn{1}{l}{\textbf{Dice (\%)}} & \multicolumn{1}{l}{\textbf{IoU (\%)}} & \multicolumn{1}{l}{\textbf{HD95 (mm)}} &   & \multicolumn{1}{l}{\textbf{Dice (\%)}} & \multicolumn{1}{l}{\textbf{IoU (\%)}} & \multicolumn{1}{l}{\textbf{HD95 (mm)}} \\
    \cmidrule{1-5}\cmidrule{7-9}\cmidrule{11-13}
    \multicolumn{2}{l}{Feddg} &   
       & 10\% 
       & 0 &   
       & $43.20 \pm 16.05^{*}$
       & $35.84 \pm 13.60^{*}$
       & $35.95 \pm 26.35^{*}$
       &   
       & $40.68 \pm 17.58^{*}$
       & $29.00 \pm 16.24^{*}$
       & $27.92 \pm 23.06^{*}$ \\     
    \multicolumn{2}{l}{Sadn} &
    &   &   &   
    & $56.90 \pm 10.25^{*}$
    & $45.51 \pm 14.29^{*}$
    & $21.07 \pm 20.11^{*}$
    &   
    & $58.06 \pm 10.45^{*}$
    & $46.40 \pm 8.91^{*}$
    & $20.12 \pm 22.18^{*}$\\
    \multicolumn{2}{l}{SLAug} &   
    &   &   &   
    & $57.54 \pm 8.09^{*}$
    & $48.47 \pm 15.07^{*}$
    & $18.99 \pm 11.40^{*}$
    &   
    & $58.00 \pm 8.36^{*}$
    & $50.05 \pm 11.74^{*}$
    & $24.53 \pm 19.81^{*}$ \\
    \multicolumn{2}{l}{LRSR} &   
    &   &   &
    & $55.15 \pm 7.09^{*}$
    & $48.82 \pm 10.16^{*}$
    & $18.27 \pm 20.35^{*}$
    &   
    & $60.24 \pm 6.47^{*}$
    & $53.80 \pm 12.09^{*}$
    & $15.77 \pm 13.26^{*}$\\
    \multicolumn{2}{l}{EPL}&  
    & 10\%  & 90\%  &   
    & $64.64 \pm 8.18^{*}$
    & $52.75 \pm 11.50^{*}$
    & $16.90 \pm 22.60^{*}$
    &   
    & $65.47 \pm 7.98^{*}$
    & $57.62 \pm 7.55^{*}$
    & $12.08 \pm 14.01^{*}$ \\  
    \multicolumn{2}{l}{MIDSS}&  
    &   &   &   
    & $57.81 \pm 15.22^{*}$
    & $45.92 \pm 17.78^{*}$
    & $23.04 \pm 28.05^{*}$
    &   
    & $59.12 \pm 10.70^{*}$
    & $50.78 \pm 14.25^{*}$
    & $18.52 \pm 15.33^{*}$ \\
    \multicolumn{2}{l}{FSDA-DG}&  
    &   &   &   
    & $66.89 \pm 11.32$
    & $58.46 \pm 11.45$
    & $15.20 \pm 10.69$
    &   
    & $77.45 \pm 7.27$
    & $66.52 \pm 8.24$
    & $15.04 \pm 13.15$ \\    
\cmidrule{1-5}\cmidrule{7-9}\cmidrule{11-13}
     \multicolumn{2}{l}{Feddg} &  
     & 20\% & 0 &   
     & $54.04 \pm 11.62^{*}$
     & $46.86 \pm 12.87^{*}$
     & $26.07 \pm 15.73^{*}$
     &   
     & $48.17 \pm 9.60^{*}$
     & $40.25 \pm 11.52^{*}$
     & $22.10 \pm 23.22^{*}$ \\    
    \multicolumn{2}{l}{Sadn} &
    &   &   &   
    & $69.29 \pm 10.38^{*}$
    & $60.21 \pm 11.05^{*}$
    & $16.28 \pm 22.05^{*}$
    &   
    & $61.32 \pm 9.17^{*}$
    & $50.85 \pm 10.57^{*}$
    & $15.50 \pm 18.68^{*}$\\ 
    \multicolumn{2}{l}{SLAug} &   
    &   &   &   
    & $70.79 \pm 8.06^{*}$
    & $62.48 \pm 13.00^{*}$
    & $16.12 \pm 9.82^{*}$
    &   
    & $61.88 \pm 9.47^{*}$
    & $50.26 \pm 11.09^{*}$
    & $15.90 \pm 9.28^{*}$ \\
    \multicolumn{2}{l}{LRSR} &   
    &   &   &
    & $70.28 \pm 10.42^{*}$
    & $63.76 \pm 14.57^{*}$
    & $15.89 \pm 12.01^{*}$
    &   
    & $67.58 \pm 13.54^{*}$
    & $56.99 \pm 12.25^{*}$
    & $13.56 \pm 11.49^{*}$\\
    \multicolumn{2}{l}{EPL}&  
    & 20\%  & 80\%  &   
    & $71.24 \pm 11.16^{*}$
    & $65.91 \pm 10.27^{*}$
    & $15.47 \pm 11.08^{*}$
    &   
    & $73.79 \pm 8.00^{*}$
    & $65.42 \pm 10.23^{*}$
    & $10.89 \pm 9.45^{*}$ \\
    \multicolumn{2}{l}{MIDSS}&  
    &   &   &   
    & $69.08 \pm 12.49^{*}$
    & $60.52 \pm 16.05^{*}$
    & $13.06 \pm 11.25^{*}$
    &   
    & $69.08 \pm 8.75^{*}$
    & $59.47 \pm 10.06^{*}$
    & $11.36 \pm 12.96^{*}$ \\
    \multicolumn{2}{l}{FSDA-DG}&  
    &   &   &   
    & $76.98 \pm 7.07$
    & $70.07 \pm 9.29$
    & $15.15 \pm 8.00$
    &   
    & $83.15 \pm 5.27$
    & $73.68 \pm 7.39$
    & $9.42 \pm 11.06$ \\
    \cmidrule{1-5}\cmidrule{7-9}\cmidrule{11-13}
    \multicolumn{2}{l}{Feddg} &  
     & 50\% & 0 &   
     & $73.01 \pm 10.97^{*}$
     & $65.46 \pm 14.24^{*}$ 
     & $19.27 \pm 14.01^{*}$
     &   
     & $65.18 \pm 10.25^{*}$
     & $55.49 \pm 11.31^{*}$
     & $15.28 \pm 13.20^{*}$ \\
    \multicolumn{2}{l}{Sadn} &
    &   &   &   
    & $82.20 \pm 13.68^{*}$
    & $65.12 \pm 9.22^{*}$
    & $19.90 \pm 20.04^{*}$
    &   
    & $67.22 \pm 12.70^{*}$
    & $58.83 \pm 11.45^{*}$
    & $13.97 \pm 13.04^{*}$\\
    \multicolumn{2}{l}{SLAug} &   
    &   &   &   
    & $82.51 \pm 9.27^{*}$
    & $66.38 \pm 13.28^{*}$
    & $18.60 \pm 9.65^{*}$
    &   
    & $74.47 \pm 9.46^{*}$
    & $60.14 \pm 8.92^{*}$
    & $14.80 \pm 7.64^{*}$ \\
    \multicolumn{2}{l}{LRSR} &   
    &   &   &
    & $81.81 \pm 8.60^{*}$
    & $66.99 \pm 10.25^{*}$
    & $15.45 \pm 11.03^{*}$
    &   
    & $75.89 \pm 8.61^{*}$
    & $63.72 \pm 11.05^{*}$
    & $12.27 \pm 8.66^{*}$\\ 
    \multicolumn{2}{l}{EPL}&  
    & 50\%  & 50\%  &   
    & $74.08 \pm 14.34^{*}$
    & $67.29 \pm 14.20^{*}$
    & $15.01 \pm 12.07^{*}$
    &   
    & $75.81 \pm 9.47^{*}$
    & $66.60 \pm 12.30^{*}$
    & $13.69 \pm 10.84^{*}$ \\  
    \multicolumn{2}{l}{MIDSS}&  
    &   &   &   
    & $81.82 \pm 15.95^{*}$
    & $74.54 \pm 11.30$
    & $12.01 \pm 11.27^{*}$
    &   
    & $76.93 \pm 8.89^{*}$
    & $71.11 \pm 10.60^{*}$
    & $8.59 \pm 8.01$ \\
    \multicolumn{2}{l}{FSDA-DG}&  
    &   &   &   
    & $84.97 \pm 7.00$
    & $74.38 \pm 9.27$
    & $10.89 \pm 7.64$
    &   
    & $85.33 \pm 6.88$
    & $76.82 \pm 6.51$
    & $8.25 \pm 7.44$ \\
    \bottomrule
    \end{tabular}%
  \end{adjustbox}
  \label{Tab4}%
\end{table*}%

%revise

\subsection{Deep Mutual Learning Strategy}
FSDA-DG incorporates a deep mutual learning strategy to facilitate collaboration among its multiple decoders. This strategy promotes knowledge exchange between decoders during training by encouraging them to learn from each other's predictions. This approach helps reduce the bias of individual decoders and mitigates the training instability that can arise from sparse supervision in a label-scarce setting.In FSDA-DG, each decoder is therefore optimized not only on its primary task loss but also to align its predictions with those of its peers. To achieve this, we define a mutual learning loss $L_{wsm}$, which takes $p\left(y_{output} \middle| x, \varepsilon_m\right)$ as input and implements a weight-sharing mechanism across branches using the Mean Squared Error (MSE) loss:

\begin{equation}
\small
\ L_{wsm}=\frac{1}{K} \frac{1}{N} \sum_{k=1,k\neq j}^{K}  {\sum_{i=1}^{N}{{(p\left(y_i\middle| x,\varepsilon_j\right)} -{p\left(y_i\middle| x,\varepsilon_k\right))}}^2}, \
\label{eq17}
\end{equation}
where $K$ represents the total number of branches, and $p\left(y_i \middle| x, \varepsilon_k\right)$ denotes the probability predicted by branch $M_k$ for pixel $i$ belonging to the target organ.

\subsection{Overall Objective Loss Function}
To define the overall loss function for self-supervised learning, we combine the consistency loss based on uncertainty estimation with the mutual learning strategy. FSDA-DG employs a weighted combination of fully supervised and unsupervised learning components, as described in Eq. \ref{eq19}. $L_{sup}$ is the Dice loss, while $\mu$ is a trade-off parameter, balancing the contributions of the supervised and unsupervised components.

During the initial stages of SSL training, the model predictions on the unlabeled data are often erroneous, and these errors are difficult to rectify in the subsequent steps. As training progresses, these errors propagate and amplify, ultimately degrading the model's performance. To address this, we use a gradual weighting mechanism for $L_{unsup}$, defined as $w_h = e^{(-0.5 \times (1-h/H)^2)}$, where $h$ denotes the current epoch number and $H$ represents the total number of training epochs (in some cases, $H$ is set to a fixed value, as suggested in \cite{wu2022mutual}). This mechanism ensures that supervised learning dominates during the early epochs, enabling the model to establish a strong foundation using labeled data. As training progresses, the weight of unsupervised learning loss ($L_{unsup}$) increases, allowing the model to learn from unlabeled data while minimizing the risk of propagating early-stage errors.

\begin{equation}
\ L_{unsup}=\gamma L_{uec}+\left(1-\gamma\right)L_{wsm}, \
\label{eq18}
\end{equation}

\begin{equation}
\ L_{total}=\mu L_{sup}+(1-\mu)w_{h}L_{unsup}. \
\label{eq19}
\end{equation}

\begin{figure}[t]
\centering
\includegraphics[width=0.45\textwidth]{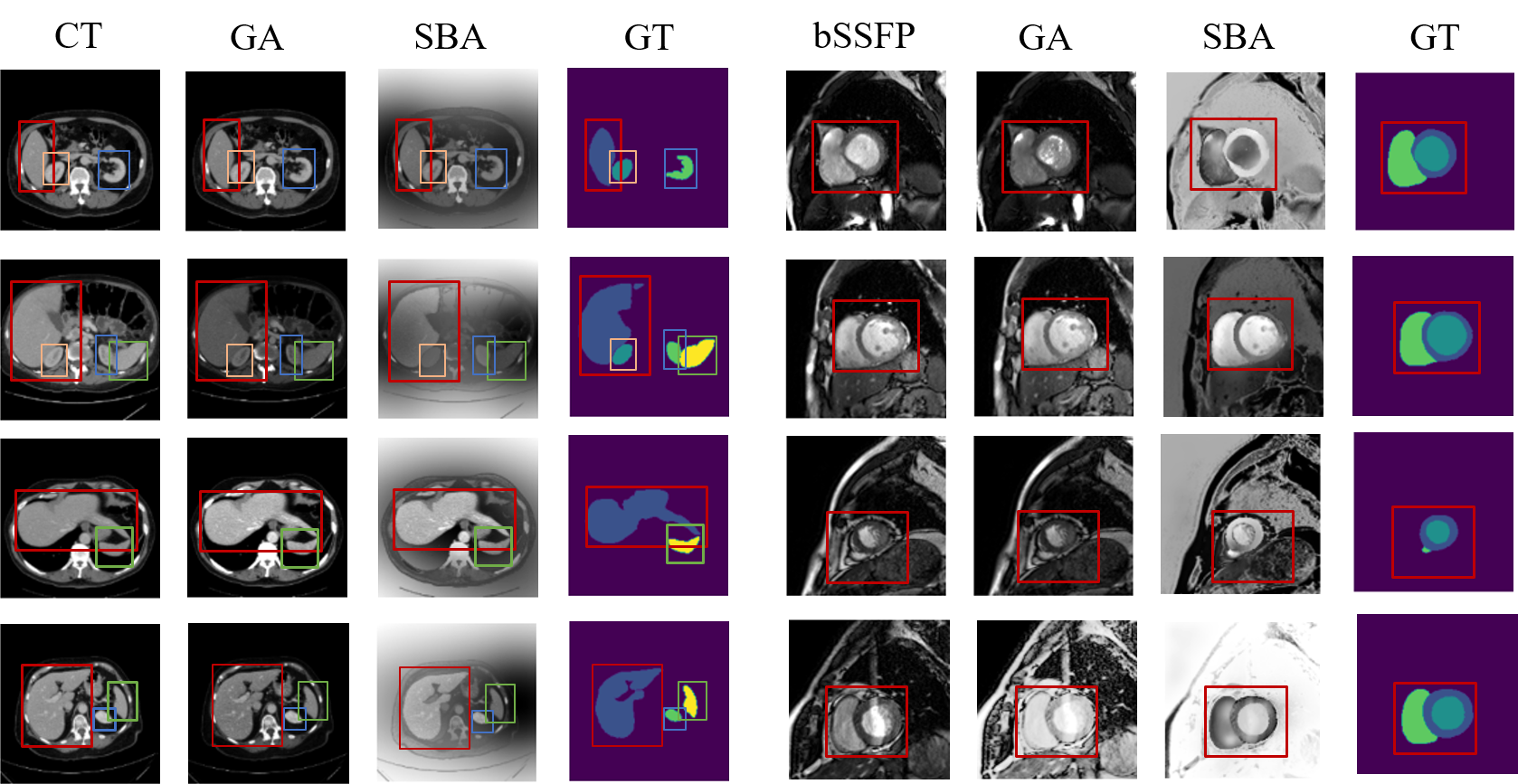} % Reduce the figure size so that it is slightly narrower than the column.
\caption{Visualization results of GA, SBA data augmentation.}
\label{add_fig2-1}
\end{figure}

\begin{figure*}[t]
\centering
\includegraphics[width=1\textwidth]{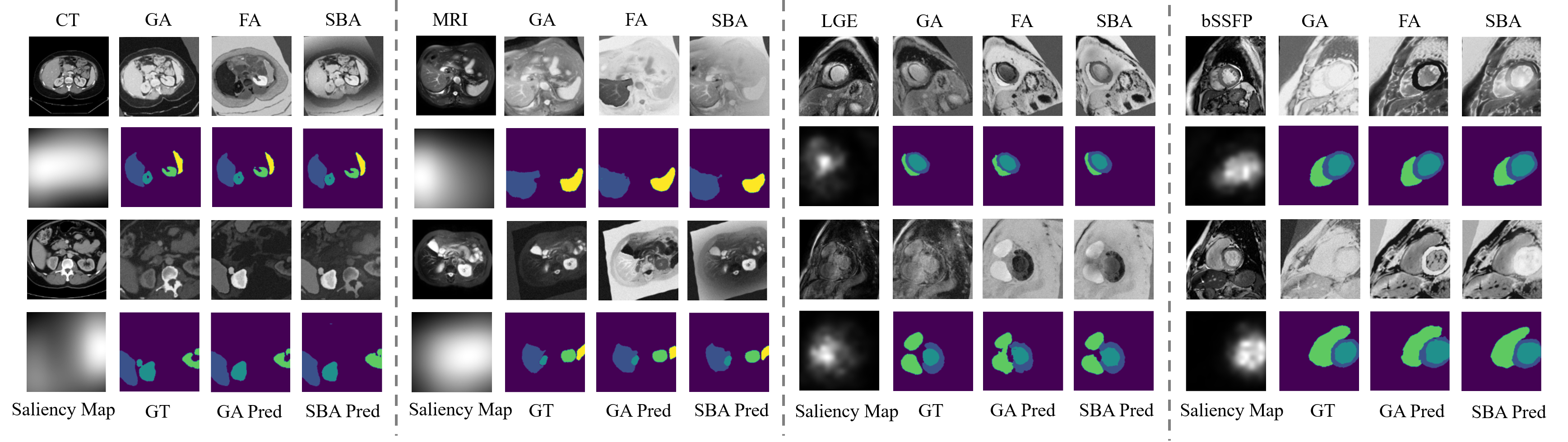} 
\caption{Visualization results of GA, FA, and SBA, along with their corresponding segmentation outputs. Note that both GA, FA, and SBA incorporate weak augmentation strategies.}
\label{fig10}
\vspace{-0.5cm}  
\end{figure*}

\section{Experiment and Results}
\subsection{Data and Processing}
The performance of FSDA-DG was evaluated using cross-modality abdominal datasets \cite{kavur2021chaos,landman2015miccai} and cross-sequence cardiac datasets \cite{zhuang2022cardiac}. The dataset division and preprocessing methods adhere to the procedures outlined in \cite{su2023rethinking}. During preprocessing, weak augmentation strategies are applied, including affine transformations, elastic deformations, brightness adjustments, contrast adjustments, gamma correction, and additive Gaussian noise. Following these standard augmentations, our proposed semantics-guided semi-supervised data augmentation is applied as an additional stage.

To ensure a fair comparison, all competing methods employ the same weak augmentation strategies. Notably, prior research has primarily focused on either SSL or DG in isolation, with limited exploration of holistic approaches that synergistically integrate both paradigms. Accordingly, to benchmark FSDA-DG, we perform independent comparisons against methods designed specifically for SSL, DG, and Semi-Supervised Learning with Out-of-Distribution Generalization (SSLODG).

\begin{figure}[t]
\centering
\includegraphics[width=0.5\textwidth]{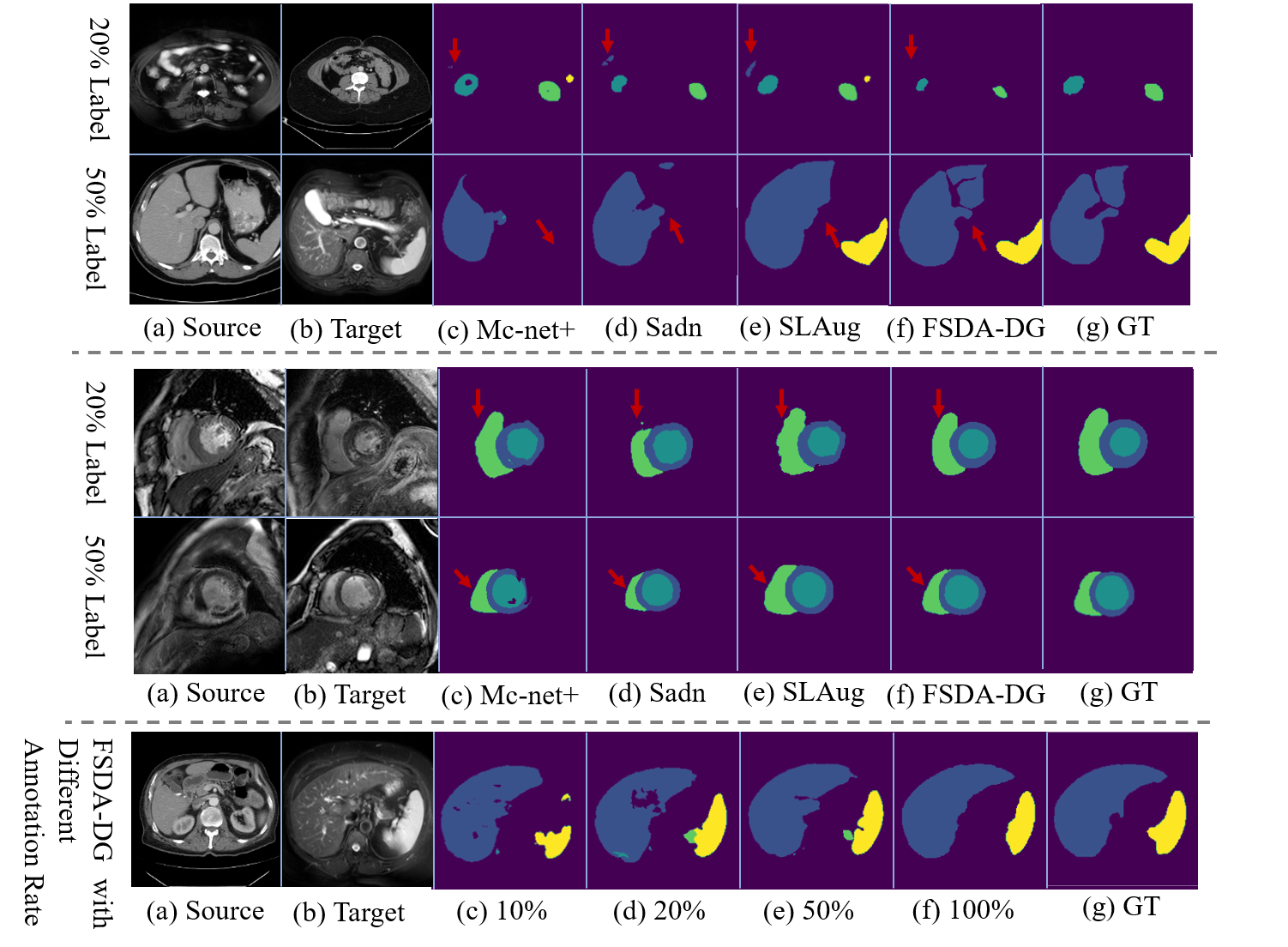} % Reduce the figure size so that it is slightly narrower than the column.
\caption{Qualitative comparisons of SSL, DG, and FSDA-DG on the target domains MRI (top two rows) and LGE (middle two rows), as well as FSDA-DG results under different labeling rates (bottom row).}
\label{fig4}
\end{figure}

\subsection{Implementation Details}
The segmentation network is based on the U-Net architecture and is trained from scratch. The grid sizes for the saliency map were empirically set to 3 and 18 for the abdominal and cardiac datasets, respectively. The model is trained using a batch size of 16, which comprises an equal number of labeled and unlabeled image patches. Optimization was performed using the Adam optimizer with an initial learning rate of $5 \times 10^{-3}$ and a weight decay of $3 \times 10^{-5}$. All experiments were conducted in PyTorch on a workstation equipped with an NVIDIA GeForce RTX 3080 GPU (12 GB VRAM).

\begin{figure*}[t]
\centering
\includegraphics[width=1\textwidth]{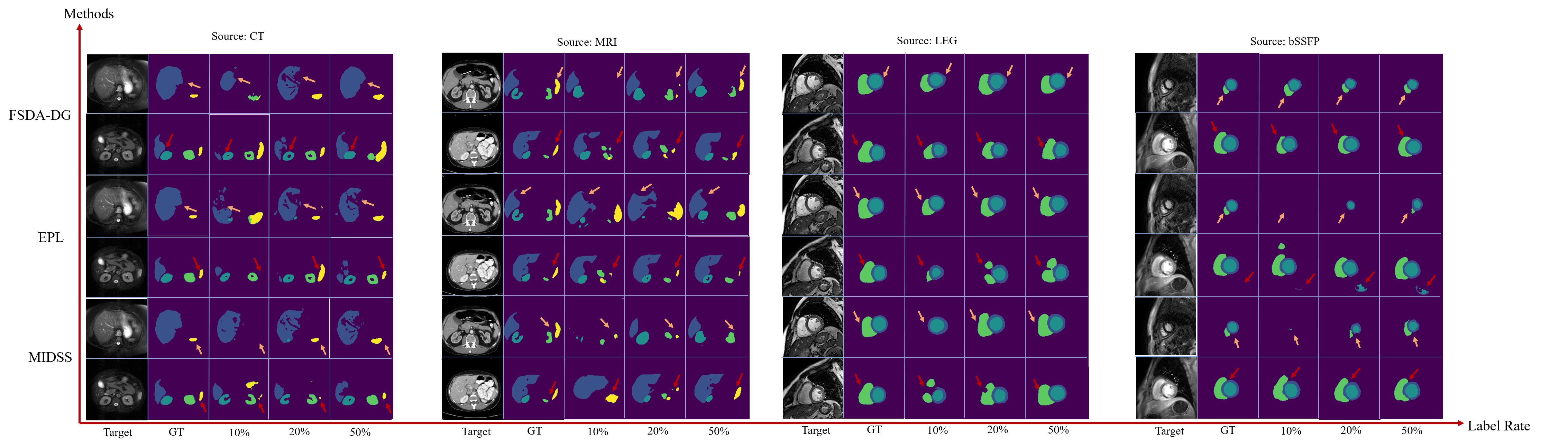} % Reduce the figure size so that it is slightly narrower than the column.
\caption{Qualitative comparison between SSLODG and FSDA-DG across various target domains and under different labeling rates.}
\label{fig5}
\end{figure*}

\subsection{Results and Comparative Analysis}
\subsubsection{Quantitative Comparison}
We evaluate FSDA-DG against state-of-the-art methods in DG, SSL, and SSLODG. The competing methods include Feddg \cite{liu2021feddg}, SADN \cite{zhou2022generalizable}, SLAug \cite{su2023rethinking}, LRSR \cite{chen2024learning}, SASSNet \cite{li2020shape}, DTC \cite{luo2021semi}, MC-Net+ \cite{wu2022mutual}, MLPRL \cite{su2024mutual},  EPL \cite{yao2022enhancing}, and MIDSS \cite{ma2024constructing}. To assess the robustness and generalization capabilities of FSDA-DG, these methods are evaluated under various labeled data configurations in both cross-modality and cross-sequence segmentation tasks. For a fair comparison, the weak augmentation strategies are uniformly applied to all methods. Additionally, we report the performance of a fully supervised U-Net model trained with varying proportions of labeled data on the target domain, which serves as the upper bound for segmentation performance. The evaluation is conducted using three metrics: Dice score, Intersection over Union (IoU), and the 95th-percentile Hausdorff distance (HD95).

The quantitative performance of FSDA-DG and all competing methods is provided in Tables~\ref{Tab1} and \ref{Tab2}, with the Dice score serving as the primary evaluation metric. Since DG methods are unable to leverage unlabeled data, no unlabeled data ($X_u = 0$) is incorporated during their training. The results reveal a significant performance gap between SSL methods and other methods. As SSL methods assume consistent data distributions between the target and source domains, they lack mechanisms for domain adaptation or generalization. Consequently, their performance deteriorates markedly in the presence of domain shifts. FSDA-DG exhibits exceptional performance under limited labeled data settings. In the cross-sequence cardiac segmentation task, FSDA-DG achieves segmentation performance comparable to LRSR (trained with 100\% labeled data), while requiring only 50\% labeled data. Notably, an interesting finding is that FSDA-DG surpasses the upper bound on the target domain LGE, achieving a higher Dice score with only 50\% labeled data. On the target domain MRI, FSDA-DG achieves substantial performance improvements by leveraging unlabeled data, increasing the Dice score from 70.93\% to 80.35\% at the 10\% labeling rate. These results highlight FSDA-DG's robustness in addressing domain shifts and its effectiveness in learning from unlabeled data, making it a highly practical approach for real-world medical image segmentation tasks.

Tables \ref{Tab3} and \ref{Tab4} provide a comprehensive evaluation of FSDA-DG compared to DG and SSLODG methods using Dice score, IoU, and HD95 as evaluation metrics. The results demonstrate that FSDA-DG achieves state-of-the-art performance in cross-modality abdominal and cross-sequence cardiac segmentation tasks. With 50\% labeled data, FSDA-DG outperforms the latest DG and SSLODG methods, LRSR and MIDSS, by 5.08\% and 5.50\% in Dice score, respectively, on the target domain MRI. Additionally, FSDA-DG achieves IoU scores of 76.03\%, 76.20\%, 74.38\%, and 76.82\% on the target domains MRI, LGE, CT, and bSSFP, respectively, significantly surpassing other DG and SSLODG methods. FSDA-DG also achieves the best HD95 performance across most target domains, demonstrating superior boundary prediction accuracy. When only 10\% of single-source domain labeled data is available, the performance of MIDSS declines significantly. This decline may be attributed to its reliance on multiple-source domain data for clipping and merging. Although clipping and merging efficiently generate data, this method introduces noise, and the generated data often fails to align with real-world data distributions. Specifically, boundary effects in the generated data result in uneven and irregular pixel distributions. With limited single-domain labeled data, MIDSS struggles to capture data distribution characteristics, leading to poor segmentation performance. As the amount of labeled data increases, MIDSS performance improves notably. These results suggest that MIDSS requires more labeled data to generalize effectively in a single-source domain setting.

Fig.~\ref{add_fig2-1} illustrates the visual outputs of the GA and SBA data augmentation strategies. Changes in anatomical structural information are reflected in variations in organ shape, boundaries, and size within the images \cite{yoon2024domain}.  The results demonstrate that our approach effectively preserves the anatomical structures of the organs to be segmented. Fig. \ref{fig10} presents the GA, FA, and SBA results of FSDA-DG, along with segmentation results at 20\% labeling rate. Saliency maps are also visualized to illustrate the model's regions of focus. These results are obtained after applying weak data augmentation. The saliency maps show that activation regions effectively cover the target segmentation sites. The results of saliency maps demonstrate superior localization performance on the LGE and bSSFP modalities, which can be attributed to the relatively concentrated regions in the cross-sequence cardiac segmentation task. In contrast, for the abdominal dataset, where target organs are more spatially distributed, the saliency maps exhibit a broader focus across the image. More importantly, the final segmentation results remain accurate across the different augmentation strategies, demonstrating FSDA-DG's ability to learn domain-invariant features and adapt to domain shifts.

\subsubsection{Visual Comparison}
Fig. \ref{fig4} presents the segmentation results of SSL and DG methods, including MC-Net+, Sadn, SLAug, and FSDA-DG. Fig. \ref{fig5} presents the segmentation results of SSLODG methods on different target domains and at varying labeling rates. These visual comparisons demonstrate that FSDA-DG accurately segments target regions of varying sizes, shapes, and spatial locations. The boundaries delineated by FSDA-DG are consistently sharper and more precise than those from competing methods, highlighting its superior boundary prediction capabilities.

\subsection{Ablation Study}
\subsubsection{Effectiveness of Semantics-Guided Semi-Supervised Data Augmentation}
We assessed the impact of the proposed data augmentation strategy on FSDA-DG at 20\% labeling rate. As shown in Tables \ref{Tab5} and \ref{Tab6}, the proposed augmentation substantially improves the baseline performance: Dice scores increased by 34.46\% and 33.83\%, IoU by 39.26\% and 41.43\%, and HD95 decreased by 9.67 mm and 13.66 mm on the target domains CT and bSSFP, respectively. These results emphasize the effectiveness of the augmentation strategy in enhancing segmentation accuracy, improving boundary precision, and facilitating generalization to unseen domains. A comparison between configuration \#3 (integrating GA and FA without SBA) and the full FSDA-DG model reveals a 3.63\% performance improvement attributable to the inclusion of SBA.This finding underscores the role of SBA in regulating the augmentation process and refining the model's ability to generalize. Overall, these results indicate that GA and SBA work synergistically, with each component playing a pivotal role in the learning of domain-invariant features.

\begin{table}[htbp]
  \centering
  \caption{Ablation study of semantics-guided semi-supervised data augmentation on the target domain CT. $\pm$ denotes the standard deviation. \(^{*}\) denotes statistical significance (\(p < 0.05\)) between the Dice score, IoU, or HD95 of a given method and that of our method, as determined by a paired t-test.}
  \begin{adjustbox}{width=0.5\textwidth}
    \begin{tabular}{lllllll}
      \toprule
      Methods & GA & FA & SBA & Dice (\%) & IoU (\%) & HD95 (mm) \\
      \midrule
        \textcolor{gray}{MC-Net+} 
        & \textcolor{gray}{\ding{55}} 
        & \textcolor{gray}{\ding{55}} 
        & \textcolor{gray}{\ding{55}} 
        & \textcolor{gray}{$40.24 \pm 7.33^{*}$}  
        & \textcolor{gray}{$29.60 \pm 11.64^{*}$} 
        & \textcolor{gray}{$28.67 \pm 15.90^{*}$} \\  
      Baseline 
      & \ding{55}
      & \ding{55} 
      & \ding{55} 
      & $42.52 \pm 7.62^{*}$ 
      & $30.81 \pm 14.11^{*}$ 
      & $24.82 \pm 18.69^{*}$ 
      \\
      \#1 
      & \ding{51} 
      & \ding{55} 
      & \ding{55} 
      & $60.94 \pm 6.82^{*}$  
      & $50.36 \pm 7.86^{*}$
      & $19.06 \pm 17.26^{*}$ \\
      \#2 
      & \ding{55} 
      & \ding{51} 
      & \ding{55} 
      & $51.98 \pm 11.00^{*}$
      & $45.62 \pm 8.05^{*}$
      & $20.08 \pm 12.23^{*}$\\
      \#3 
      & \ding{51} 
      & \ding{51} 
      & \ding{55} 
      & $73.05 \pm 9.23^{*}$ 
      & $64.60 \pm 7.64^{*}$ 
      & $16.80 \pm 13.50^{*}$\\
      \#4 
      & \ding{55} 
      & \ding{51} 
      & \ding{51} 
      & $64.99 \pm 6.67^{*}$ 
      & $58.36 \pm 9.08^{*}$ 
      & $18.12 \pm 15.59^{*}$\\
      FSDA-DG 
      & \ding{51} 
      & \ding{55} 
      & \ding{51} 
      & \textbf{$76.98 \pm 7.07$}
      & \textbf{$70.07 \pm 9.29$}
      & \textbf{$15.15 \pm 8.00$}\\
      \bottomrule
    \end{tabular}
  \end{adjustbox}
  \label{Tab5}
\end{table}

\begin{table}[htbp]
  \centering
  \caption{Ablation study of semantics-guided semi-supervised data augmentation on the target domain bSSFP, $\pm$ denotes the standard deviation. \(^{*}\) denotes statistical significance (\(p < 0.05\)) between the Dice score, IoU, or HD95 of a given method and that of our method, as determined by a paired t-test.}
  \begin{adjustbox}{width=0.5\textwidth}
    \begin{tabular}{lllllll}
      \toprule
      Methods & GA & FA & SBA & Dice (\%) & IoU (\%) & HD95 (mm) \\
      \midrule
        \textcolor{gray}{MC-Net+} 
        & \textcolor{gray}{\ding{55}} 
        & \textcolor{gray}{\ding{55}} 
        & \textcolor{gray}{\ding{55}} 
        & \textcolor{gray}{$44.55 \pm 6.26^{*}$}  
        & \textcolor{gray}{$31.54 \pm 8.90^{*}$} 
        & \textcolor{gray}{$23.09 \pm 28.19^{*}$} \\  

      Baseline 
      & \ding{55}
      & \ding{55} 
      & \ding{55} 
      & $49.32 \pm 6.29^{*}$ 
      & $32.25 \pm 12.01^{*}$ 
      & $23.08 \pm 25.30^{*}$ 
      \\
      \#1 
      & \ding{51} 
      & \ding{55} 
      & \ding{55} 
      & $67.29 \pm 3.11^{*}$  
      & $56.14 \pm 7.70^{*}$
      & $16.50 \pm 21.22^{*}$ \\
      \#2 
      & \ding{55} 
      & \ding{51} 
      & \ding{55} 
      & $70.43 \pm 4.98^{*}$
      & $60.68 \pm 5.11^{*}$
      & $17.36 \pm 20.38^{*}$\\
      \#3 
      & \ding{51} 
      & \ding{51} 
      & \ding{55} 
      & $79.80 \pm 3.55^{*}$ 
      & $72.32 \pm 7.45$ 
      & $14.90 \pm 15.68^{*}$\\
      \#4 
      & \ding{55} 
      & \ding{51} 
      & \ding{51} 
      & $81.77 \pm 6.20^{*}$ 
      & $69.99 \pm 9.05^{*}$ 
      & $13.92 \pm 15.28^{*}$\\
      FSDA-DG 
      & \ding{51} 
      & \ding{55} 
      & \ding{51} 
      & \textbf{$83.15 \pm 5.27$}
      & \textbf{$73.68 \pm 7.39$}
      & \textbf{$9.42 \pm 11.06$}\\
      \bottomrule
    \end{tabular}
  \end{adjustbox}
  \label{Tab6}
\end{table}

\subsubsection{Effectiveness of Consistency Loss based on Uncertainty Estimation and Deep Mutual Learning Strategy}
To evaluate the effectiveness of the proposed consistency loss and deep mutual learning strategy, we compare it against several alternative configurations. As shown in Tables~\ref{Tab7} and \ref{Tab8}, the integration of $L_{une}$ yields a notable improvement in Dice scores, with increases of 3.07\% on the CT target domain and 4.12\% on the bSSFP target domain compared to the baseline. Incorporating $F_{sharp}$ or $L_{wsm}$ individually provides limited improvements. The full FSDA-DG model, which synergistically integrates all components, consistently outperformed all other configurations. These results underscore the pivotal role of uncertainty evaluation in enhancing semi-supervised learning by enabling the model to focus on more reliable predictions during training.

\subsubsection{Adjustment of $T$, $ \alpha$, $\mu$ and $\gamma$}
\begin{figure}[t]
\centering
\includegraphics[width=0.5\textwidth]{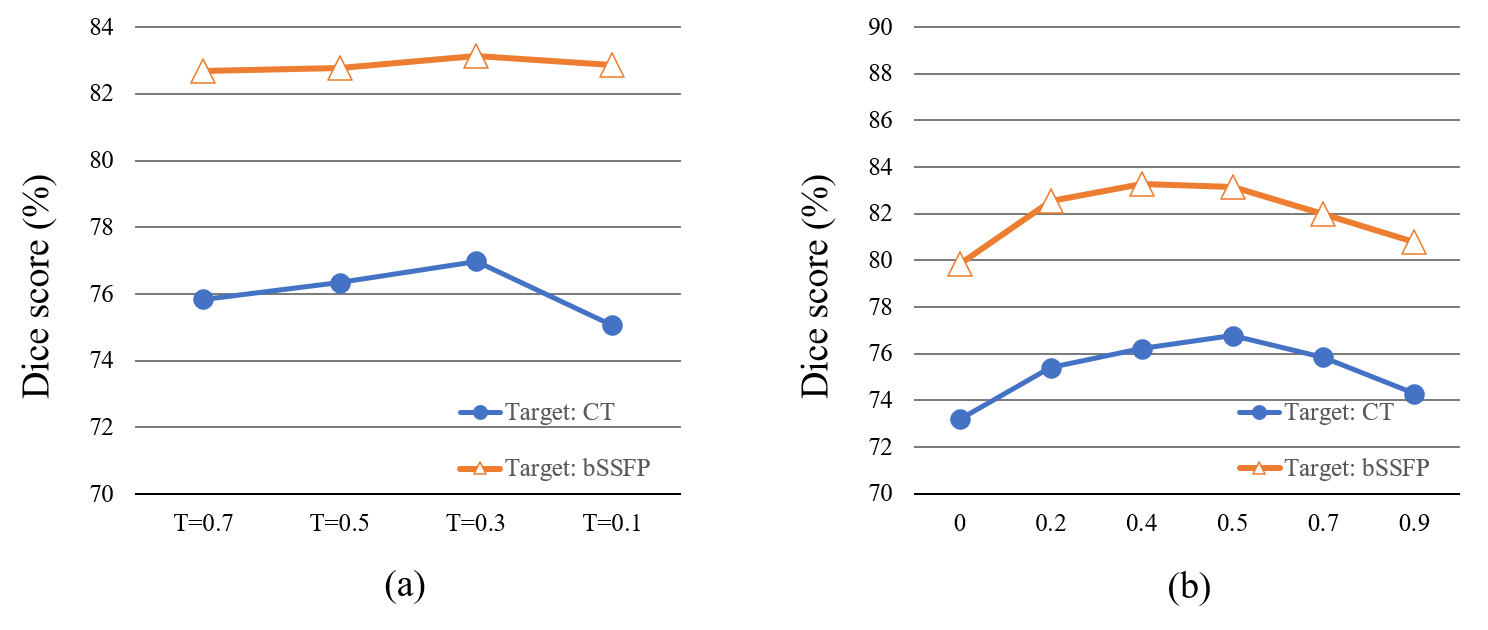} % Reduce the figure size so that it is slightly narrower than the column.
\caption{(a) Performance of FSDA-DG w.r.t. $T$. (b) Performance of FSDA-DG w.r.t. $\alpha$. }
\label{fig6}
\end{figure}

\begin{figure}[t]
\centering
\includegraphics[width=0.4\textwidth]{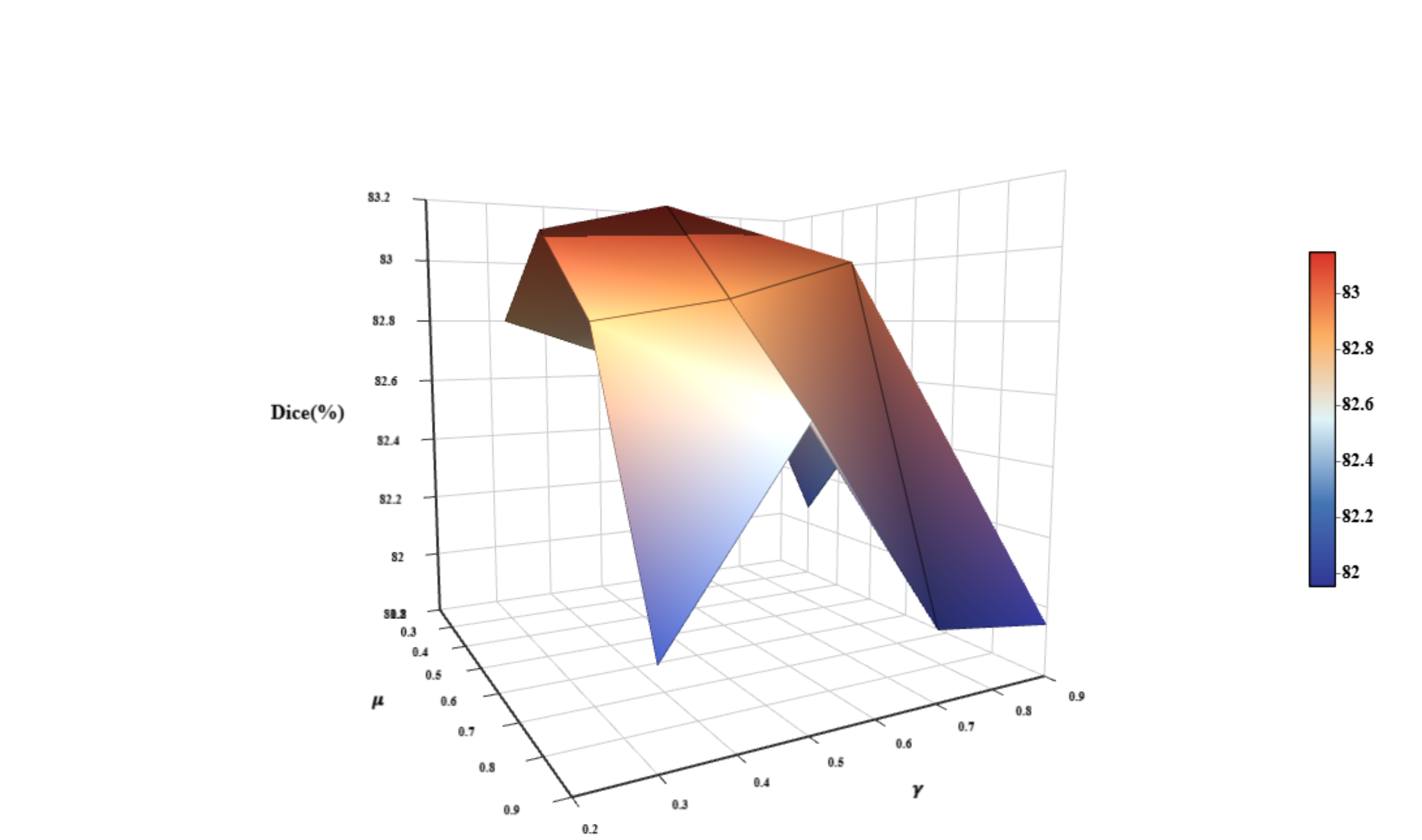} % Reduce the figure size so that it is slightly narrower than the column.
\caption{Performance of FSDA-DG w.r.t. $\mu$ and $\gamma$.}
\label{fig7}
\end{figure}

$T$ is the hyperparameter used to control the temperature of sharpening. $\alpha$, $\mu$ and $\gamma$ are employed to  balance the overall loss. As shown in Fig.~\ref{fig6} and \ref{fig7}, the results indicate that FSDA-DG achieves the best performance when $T$ is set to 0.3 and $\alpha$, $\mu$ and $\gamma$ are each set to 0.5.

\begin{table}[htbp]
  \centering
  \caption{Ablation study of SSL framework on target domain CT. $\pm$ denotes the standard deviation. \(^{*}\) denotes statistical significance (\(p < 0.05\)) between the Dice score, IoU, or HD95 of a given method and that of our method, as determined by a paired t-test.}
  \begin{adjustbox}{width=0.49\textwidth}
    \begin{tabular}{lllllll}
      \toprule
      Methods & $F_{sharp}$ & $L_{une}$ & $L_{wsm}$ & Dice (\%) & IoU (\%) & HD95 (mm)\\
      \midrule
      Baseline 
      & \ding{55} & \ding{55} & \ding{55} 
      & $71.25 \pm 9.85^{*}$  
      & $59.65 \pm 10.01^{*}$  
      & $19.30 \pm 24.34^{*}$   \\

      \#1 & \ding{51} & \ding{55} & \ding{55} 
      & $72.35 \pm 7.22^{*}$
      & $58.70 \pm 10.00^{*}$
      & $20.48 \pm 18.17^{*}$\\

      \#2 & \ding{55} & \ding{51} & \ding{55} 
      & $74.32 \pm 7.35^{*}$
      & $61.68 \pm 8.35^{*}$
      & $17.31 \pm 16.98^{*}$\\

      \#3 & \ding{55} & \ding{55} & \ding{51} 
      & $72.23 \pm 6.78^{*}$
      & $61.06 \pm 8.94^{*}$
      & $19.65 \pm 15.21^{*}$\\

      \#4 & \ding{51} & \ding{51} & \ding{55} 
      & $75.32 \pm 7.12^{*}$
      & $67.39 \pm 7.90^{*}$
      & $14.01 \pm 14.10^{*}$\\

      \#5 & \ding{55} & \ding{51} & \ding{51} 
      &$76.04 \pm 6.51$
      & $68.96 \pm 7.31^{*}$
      & $17.01 \pm 12.39^{*}$\\

      \#6 & \ding{51} & \ding{55} & \ding{51} 
      & $73.20 \pm 4.68^{*}$ 
      & $67.56 \pm 6.92^{*}$ 
      & $17.43 \pm 13.22^{*}$ \\
      
      FSDA-DG & \ding{51} & \ding{51} & \ding{51} 
      & 76.98$ \pm 7.07 $ 
      & 70.17$ \pm 9.29$ 
      & 15.15$ \pm 8.00$ \\
      \bottomrule
    \end{tabular}
  \end{adjustbox}
  \label{Tab7}
\end{table}

\begin{table}[htbp]
  \centering
  \caption{Ablation study of SSL framework on target domain bSSFP. $\pm$ denotes the standard deviation. \(^{*}\) denotes statistical significance (\(p < 0.05\)) between the Dice score, IoU, or HD95 of a given method and that of our method, as determined by a paired t-test.}
  \begin{adjustbox}{width=0.49\textwidth}
    \begin{tabular}{lllllll}
      \toprule
      Methods & $F_{sharp}$ & $L_{une}$ & $L_{wsm}$  & Dice (\%) & IoU (\%) & HD95 (mm) \\
      \midrule
      Baseline 
      & \ding{55} & \ding{55} & \ding{55} 
      & $78.90 \pm 6.31^{*}$  
      & $70.83 \pm 7.01^{*}$  
      & $19.30 \pm 21.00^{*}$   \\

      \#1 & \ding{51} & \ding{55} & \ding{55} 
      & $79.11 \pm 7.15^{*}$
      & $71.60 \pm 5.31^{*}$
      & $21.48 \pm 13.48^{*}$\\

      \#2 & \ding{55} & \ding{51} & \ding{55} 
      & $83.02 \pm 9.66^{*}$
      & $72.60 \pm 7.37^{*}$
      & $14.24 \pm 15.39^{*}$\\

      \#3 & \ding{55} & \ding{55} & \ding{51} 
      & $79.65 \pm 6.15^{*}$
      & $70.00 \pm 6.38^{*}$
      & $17.26 \pm 10.62^{*}$\\

      \#4 & \ding{51} & \ding{51} & \ding{55} 
      & $82.73 \pm 5.17$
      & $72.62 \pm 9.02^{*}$
      & $15.76 \pm 12.52^{*}$\\

      \#5 & \ding{55} & \ding{51} & \ding{51} 
      &$82.56 \pm 5.07$
      & $72.76 \pm 8.14$
      & $14.01 \pm 14.26^{*}$\\

      \#6 & \ding{51} & \ding{55} & \ding{51} 
      & $79.80 \pm 4.68^{*}$ 
      & $68.55 \pm 9.09^{*}$ 
      & $13.26 \pm 16.91^{*}$ \\
      
      FSDA-DG & \ding{51} & \ding{51} & \ding{51} 
      & 83.15$ \pm 5.27 $ 
      & 73.68$ \pm 7.39$ 
      & 9.42$ \pm 11.06$ \\
      \bottomrule
    \end{tabular}
  \end{adjustbox}
  \label{Tab8}
\end{table}

% \begin{table}[htbp]
%   \centering
%   \caption{Ablation study of SSL framework. Dice score is used as the evaluation metric.}
%   \begin{adjustbox}{width=0.49\textwidth}
%     \begin{tabular}{cccccc}
%       \toprule
%       Methods & $F_{sharp}$ & $L_{une}$ & $L_{wsm}$ & MRI-CT & LGE-bSSFP \\
%       \midrule
%       Baseline & \ding{55} & \ding{55} & \ding{55} & 71.25 & 78.90 \\
%       \#1 & \ding{51} & \ding{55} & \ding{55} & 72.35 & 79.11 \\
%       \#2 & \ding{55} & \ding{51} & \ding{55} & 74.32 & 83.02 \\
%       \#3 & \ding{55} & \ding{55} & \ding{51} & 72.23 & 79.65 \\
%       \#4 & \ding{51} & \ding{51} & \ding{55} & 75.32 & 82.73 \\
%       \#5 & \ding{55} & \ding{51} & \ding{51} & 76.04 & 82.56 \\
%       \#6 & \ding{51} & \ding{55} & \ding{51} & 73.20 & 79.85 \\
%       FSDA-DG & \ding{51} & \ding{51} & \ding{51} & \textbf{76.98} & \textbf{83.15} \\
%       \bottomrule
%     \end{tabular}
%   \end{adjustbox}
%   \label{Tab4}
% \end{table}

\subsubsection{Effectiveness of Multiple Decoders and Outputs}
FSDA-DG employs transposed convolutional, linear interpolation, and nearest interpolation layers to construct distinct decoders. The introduction of multiple branches inevitably increases computational overhead during both training and inference. As shown in Table~\ref{Tab9}, adding decoder branches does not lead to notable improvements in segmentation accuracy.Consequently, a configuration with three decoder branches was selected to balance model diversity and computational cost. Furthermore, ensembling the outputs from the multiple decoders at inference time did not improve segmentation accuracy, as the individual decoders converged to similar solutions after training. To minimize inference costs, we selected the original encoder-decoder architecture, specifically the shared encoder and the first decoder, as the final model for testing.

\begin{table}[htbp]
  \centering
  \caption{Ablation study of multiple decoders and outputs on the target domain LGE at a 20\% labeling rate. The metrics include Dice score, IoU, and HD95. A single output refers to using only the result from the original encoder-decoder architecture, while multiple outputs involve summing the results from all encoder-decoder architectures. Model complexities, such as the number of parameters (Para.) and multiply-accumulate operations (MACs), are measured during model inference. $\pm$ denotes the standard deviation. \(^{*}\) denotes statistical significance (\(p < 0.05\)) between the Dice score, IoU, or HD95 of a given method and that of our method, as determined by a paired t-test.}
    \begin{adjustbox}{width=0.5\textwidth}
    \begin{tabular}{lllllll}
      \toprule
      Methods & Output & Dice (\%) & IoU (\%) & HD95 (mm) &Para.(M)  &MACs(G) \\
      \midrule
      \textcolor{gray}{U-Net }                            &\textcolor{gray}{ - }       &\textcolor{gray}{ -  }        &\textcolor{gray}{ - }      &\textcolor{gray}{ - }      & \textcolor{gray}{1.81}  & \textcolor{gray}{2.99}  \\
      2-Decoders & Single     
      & $82.17 \pm 6.08$   
      &  $69.30 \pm 7.91^{*}$    
      &   $14.98 \pm 9.58^{*}$    
      & 1.81 & 2.99 \\
      2-Decoders & Multiple   
      & $81.98 \pm 5.79^{*}$   
      &   $69.49 \pm 7.46^{*}$  
      &      $14.02 \pm 9.52^{*}$   
      &  2.23 
      &  4.05 \\
      3-Decoders & Single    
      &    $84.56 \pm 6.76$   
      &     $74.29 \pm 8.06$   
      &       $11.11  \pm 10.09$  
      &  1.81  
      & 2.99 \\
      3-Decoders & Multiple  
      &    $84.35 \pm 6.64$
      &      $73.99 \pm 8.91$
      &       $13.37 \pm 12.40^{*}$
      &  2.58  
      &  5.17\\
      4-Decoders & Single     
      &    $84.69 \pm 6.87$
      &      $74.41 \pm 7.33$
      &       $11.74  \pm 9.35$  
      &  1.81  
      & 2.99 \\
      4-Decoders & Multiple   
      &     $84.74 \pm 5.93$
      &      $74.86 \pm 7.66$
      &    $10.79  \pm 7.12^{*}$ 
      &   3.62 
      & 7.35 \\
      \bottomrule
    \end{tabular}
      \end{adjustbox}
  \label{Tab9}
\end{table}

\begin{figure}[t]
\centering
\includegraphics[width=0.46\textwidth]{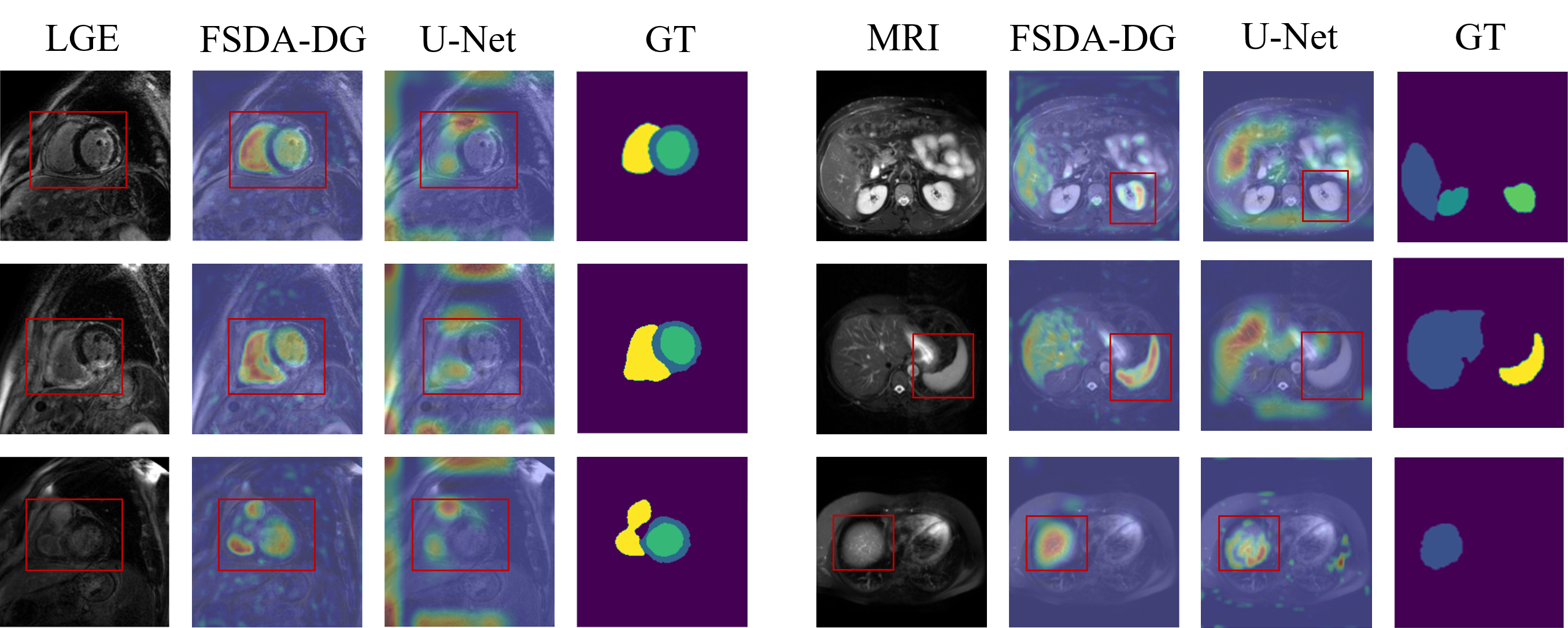} % Reduce the figure size so that it is slightly narrower than the column.
\caption{Visualization results of feature activation regions in the final convolutional layer of the encoder for FSDA-DG and U-Net, on the target domain LGE and MRI.}
\label{add_fig2-2}
\end{figure}

\section{Discussions}
\subsection{Guide for Network Selection}
FSDA-DG is based on the pure CNN architecture, whereas the Transformer-based architecture represents an alternative and effective paradigm for medical image segmentation. This section compares the performance of these two architectural paradigms and discusses the factors contributing to the observed differences.

\begin{table}[htbp]
\centering
\caption{Performance comparison between FSDA-DG (U-Net) and transformer-based networks (SwinUNet and TransUNet) on the target domain CT and bSSFP  at 20\% labeling rate.}
\begin{adjustbox}{width=0.5\textwidth}
\begin{tabular}{lllll}
\toprule
Network & Target Domain & Dice (\%) & IoU (\%) & HD95 (mm) \\ 
\midrule
U-Net        & CT                     & 76.98 ± 7.07       & 70.07 ± 9.29      & 15.15 ± 8.00       \\ 
SwinUNet     & -                     & 74.01 ± 6.30       & 66.46 ± 11.28     & 15.96 ± 13.12      \\ 
TransUNet    & -                    & 73.54 ± 7.60       & 64.01 ± 8.25      & 18.45 ± 21.38      \\ 
U-Net        & bSSFP                  & 83.15 ± 5.27       & 73.68 ± 7.39      & 9.42 ± 11.06       \\ 
SwinUNet    & -                 & 79.28 ± 8.33       & 69.34 ± 10.04     & 12.02 ± 14.51      \\ 
TransUNet    & -                 & 76.99 ± 7.46       & 64.72 ± 13.10     & 13.01 ± 15.26      \\ 
\bottomrule
\end{tabular}
\end{adjustbox}
\label{tab:CNN and Transformer}
\end{table}
Table~\ref{tab:CNN and Transformer} presents the segmentation performance of the different networks. FSDA-DG, based on the U-Net backbone, consistently outperformed transformer-based networks such as SwinUNet \cite{cao2022swin} and TransUNet \cite{chen2021transunet}. These results suggest that CNN-based architectures are better suited for the specific challenges of domain generalization in limited-data medical imaging scenarios. The observed performance differences can likely be attributed to the following factors:

\begin{enumerate}
    \item \textbf{Data and Supervision Requirements}: Transformers are highly dependent on large-scale datasets to realize the full potential of their global modeling capabilities. Consequently, in data-scarce and semi-supervised settings, they are more prone to overfitting and training instability. In contrast, CNNs, with their inherent inductive biases, exhibit superior sample efficiency and robustness.  

    \item \textbf{Sensitivity to Domain Shifts}: While Transformers excel at modeling global dependencies, their performance can be sensitive to the intensity variations and style changes common in the presence of domain shifts. CNNs, conversely, are more adept at preserving local anatomical consistency and are often more compatible with domain-specific augmentation strategies.  

    \item \textbf{Parameter Complexity}: Transformers typically have a significantly greater number of parameters, owing to their multi-head attention mechanisms and large patch embeddings. This higher parameter count exacerbates optimization challenges, particularly when training from scratch with limited labeled data. 
\end{enumerate}
Additionally, these results suggest that performance variations arising from architectural differences are less significant than those induced by the choice of data augmentation and training strategies. This finding highlights the critical importance of effective data processing and optimization techniques for achieving robust domain generalization. Future work should therefore continue to prioritize the development of advanced data augmentation methods and improved training strategies.

\subsection{Feature Visualization}
To further investigate the domain-invariant feature learning capability of FSDA-DG, we use GradCAM \cite{selvaraju2017grad} to visualize the activation regions in the encoder's convolutional layer. As shown in Fig.~\ref{add_fig2-2}, the visualizations reveal that even in the presence of a domain shift, FSDA-DG produces activation regions that are well-aligned with the actual organ locations. In contrast, conventional segmentation models exhibit multiple, widely dispersed activation areas. This observation aligns with recent findings \cite{litowards} and demonstrates that FSDA-DG successfully identifies class-discriminative, domain-invariant features. In comparison, a conventional segmentation model, as explained by the "lottery ticket hypothesis," tends to capture only a subset of these features.

\subsection{Limitations and Future Work}
FSDA-DG addresses a novel and underexplored setting of single-source domain generalization with limited annotations. While our work demonstrates promising results, several limitations merit further investigation. First, the data augmentation strategy, though effective in preserving organ-level semantics, relies on heuristic design. Future work could explore automated or learning-based approaches to enable more adaptive and robust transformations. Second, FSDA-DG provides valuable empirical insights into domain-invariant learning. However, a rigorous theoretical framework for domain-invariant learning, particularly in medical imaging, remains underdeveloped and necessitates further study. Lastly, while our evaluation focuses on segmentation accuracy under domain shifts, broader evaluation metrics should be incorporated in future work to enhance the reliability, interpretability, and practical utility of domain generalization methods.

\section{Conclusions}
The challenges of domain shifts and label scarcity in medical imaging have garnered significant attention, prompting the development of methods such as SSL, DG, and DA. These methods have demonstrated state-of-the-art performance in addressing specific challenges. However, in clinical scenarios, domain shifts and label scarcity frequently coexist, necessitating the design of solutions that address these interdependent issues simultaneously.

In this study, we propose FSDA-DG, a unified solution to tackle the challenges of domain shifts and label scarcity in medical imaging. FSDA-DG uses semantics-guided semi-supervised data augmentation to transform data into extensive domain knowledge while preserving domain-invariant semantic information. Moreover, by constraining different perturbed branches to produce consistent predictions, FSDA-DG facilitates domain-invariant feature learning. Extensive experimental results demonstrate that FSDA-DG reduces reliance on large labeled datasets and enhances the adaptability of segmentation models to unseen target domains. We hope this research draws greater attention to the important problem of domain-generalizable SSL and encourages the development of more principled and robust learning methodologies for medical image analysis.

\section{Acknowledgments}
This work was supported in part by the National Natural Science Foundation of China (grant numbers 62371221, 12326616, and 62201245); the National High-end Foreign Experts Recruitment Plan (grant number G2023030025L); the Science and Technology Program of Guangdong Province (grant number 2022A0505050039); Yunnan Fundamental Research Project (grant numbers 202401AU070215 and 202501AT070238); Yunnan University Medical Research Foundation (grant number YDYXJJ2024-0021); the Open Project Program of the Yunnan Key Laboratory of Intelligent Systems and Computing (grant number ISC23Y02); Zhejiang Provincial Natural Science Foundation (grant number LO23F030002).

\printcredits

\bibliographystyle{cas-model2-names}

\bibliography{cas-refs}

\end{document}